\documentclass[journal]{IEEEtran}
\usepackage[T1]{fontenc}    % use 8-bit T1 fonts
\usepackage{hyperref}       % hyperlinks
\usepackage{url}            % simple URL typesetting
\usepackage{booktabs}       % professional-quality tables
\usepackage{amsfonts}       % blackboard math symbols
\usepackage{nicefrac}       % compact symbols for 1/2, etc.
\usepackage{microtype}      % microtypography
\usepackage{lipsum}
\usepackage{fancyhdr}       % header
\usepackage{epsfig}
\usepackage{graphicx}       % graphics
\usepackage{mathtools}   % loads »amsmath«
\usepackage{amsmath}
\usepackage{amssymb}
\graphicspath{{media/}}     % organize your images and other figures under media/ folder
\usepackage{iftex}
\usepackage{float}
\usepackage{makecell}
\usepackage{xcolor}
\usepackage{lineno}
\ifTUTeX
\usepackage{fontspec}
\else
\usepackage[T1]{fontenc}
\usepackage[utf8]{inputenc} % The default since 2018
\DeclareUnicodeCharacter{200B}{{\hskip 0pt}}
\fi
\ifCLASSINFOpdf
\else
\fi
\begin{document}
		\title{On the Relation between Optical Aperture and Automotive Object Detection}
		
		\author{Ofer Bar-Shalom,
			Tzvi Philipp,
			and~Eran Kishon% <-this % stops a space
			\thanks{Authors are with General Motors R\&D Division, Software-Defined Vehicle Research (SDVR),
				General Motors Technical Center, 13 Arye Shenkar St., Herzliya, 4672513, Israel.\\
				Corresponding Author:~\texttt{oferbarshalom@gmail.com}}%, tzvi76@gmail.com} \\
				%\texttt{eran.kishon@gm.com}}% <-this % stops a space
			}
		

	\markboth{IEEE Transactions on Intelligent Vehicles,~Vol.~XX, No.~YY, Month~2025}%
	{Bar-Shalom \MakeLowercase{\textit{et al.}}:On the Relation between Optical Aperture and Automotive Object Detection}



	\maketitle
	
	\begin{abstract}
		We investigate the impact of aperture size and shape in an automotive camera imaging system on the performance of deep-learning-based object detection tasks, such as digit recognition on speed signs, traffic sign detection and classification, and traffic light state identification. This paper presents a method for simulating the optical effects of the imaging system’s aperture by emulating the point spread function (PSF), enhancing scene realism, and bridging the domain gap between synthetic and real-world images. Our study utilizes computer-generated synthetic scenes, initially created without optical distortions, which are further refined using this method.
	\end{abstract}
	

	% \fi
	\IEEEpeerreviewmaketitle
		\section{Introduction}
	%	Camera imaging systems form the predominant sensing modality used for any level of autonomous driving. Their relative low-pricing and versatile performance in a variety of lighting and weather conditions makes them the ultimate go-to solution for the automotive industry. 
	\IEEEPARstart{C}amera imaging systems are the primary sensing technology used across all levels of autonomous driving. Their affordability and adaptability to different lighting and weather conditions make them an ideal solution for the automotive industry.
	Roughly speaking, the camera imaging system may be divided into 3 sections: the optics, the aperture (or the pupil) and the imaging sensor (commonly built as a composite metal oxide/CMOS array of photo-diodes) \cite{CMOSsensors, Yadid}.
	
	In the following paper the relation between the camera pupil shape and size and its effect on the perception system in terms of object detection is studied. The original idea was inspired by the animal kingdom; animal pupils come in diverse shapes, each adapted to specific environmental needs. 
	\color{black}Pupil shape is closely linked to an animal’s ecological role and visual needs, as explored by Banks \textit{et al.}\cite{Banks}. Horizontal, rectangular pupils, common in grazing herbivores like goats, maximize panoramic vision, helping detect predators across wide terrains while grazing. These pupils align with the horizon, providing stable vision even when the head tilts downward. Conversely, vertical slit pupils, typical of ambush predators, enhance depth perception and focus, aiding in distance estimation for striking prey. Circular pupils are seen in active foragers, optimizing light capture across various environments and activities. 
	
	\color{black}
	In the context of the automotive world, a natural question emerges: is there a particular aperture (pupil) shape that outperforms the conventional fixed-size circular aperture in automotive object detection tasks?
	As will be shown, the answer may be not straightforward.
	
\begin{figure}[!ht]
		\centering
		%		\fbox{\rule[-.5cm]{4cm}{4cm} \rule[-.5cm]{4cm}{0cm}}
		\resizebox{3.5in}{!}{\includegraphics{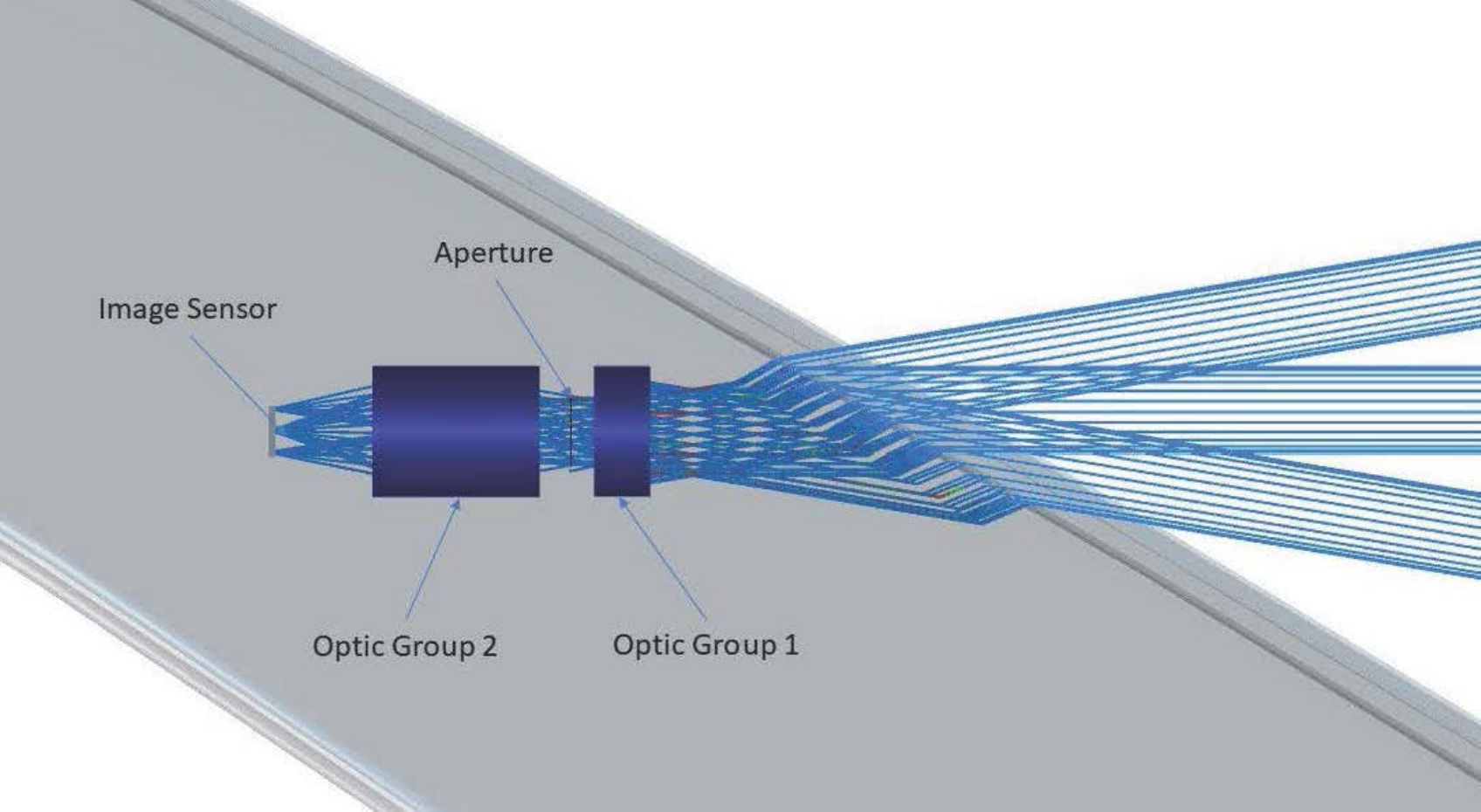}}
		\caption{Imaging \& Optical System Model}
		\label{fig11}
	\end{figure}

\begin{figure}[!ht]
		\centering
		\resizebox{3.0in}{!}{\includegraphics{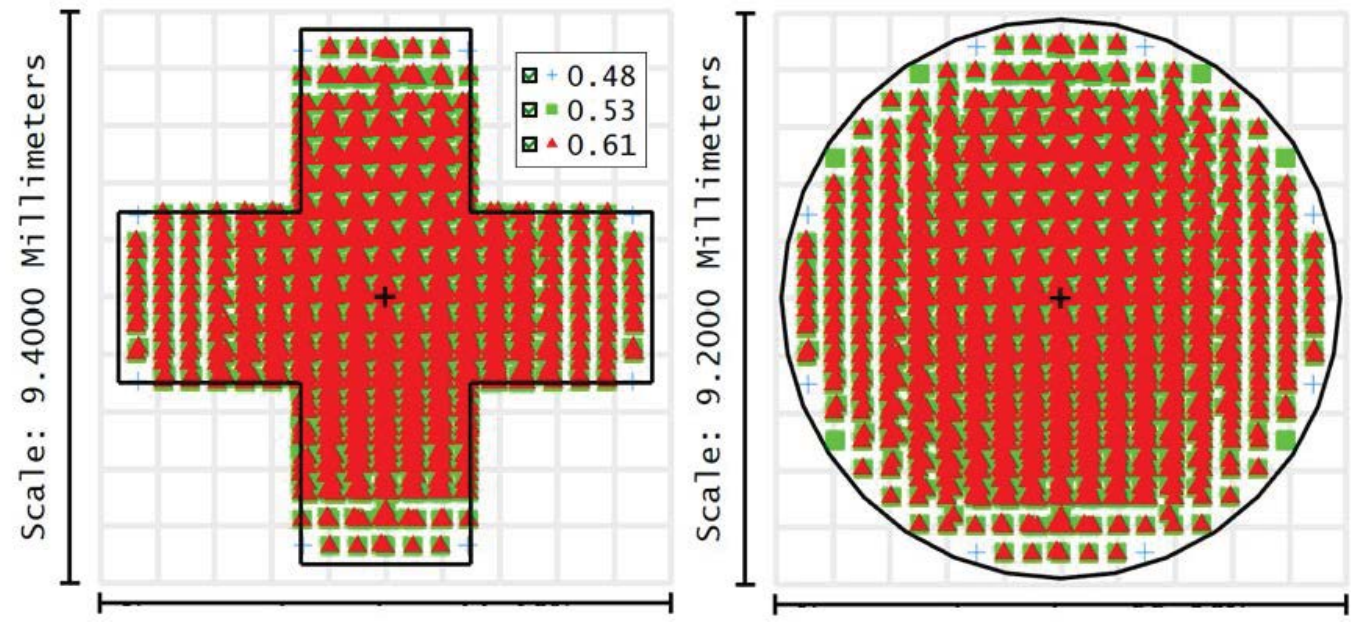}}
		\caption{`Plus' \& Circular-Shaped Apertures}
		\label{fig1}
	\end{figure}

	Figure~\ref{fig11} illustrates the imaging system, showing a side view of the camera mounted behind the windshield of a General Motors Cadillac Escalade. This system can utilize different aperture designs, such as a conventional circular aperture commonly used in automotive imaging systems, a non-conventional plus-shaped aperture or any other shapes. Ray Tracing analysis for these sample apertures, performed using Zemax$^{\textrm{TM}}$, is shown in Fig.~\ref{fig1} for three representative wavelengths: 480 nm, 530 nm, and 610 nm. For both apertures some lateral color aberrations can be observed.
	
\begin{figure}[!ht]
		\centering
		%		\fbox{\rule[-.5cm]{4cm}{4cm} \rule[-.5cm]{4cm}{0cm}}
		\resizebox{3.5in}{!}{\includegraphics{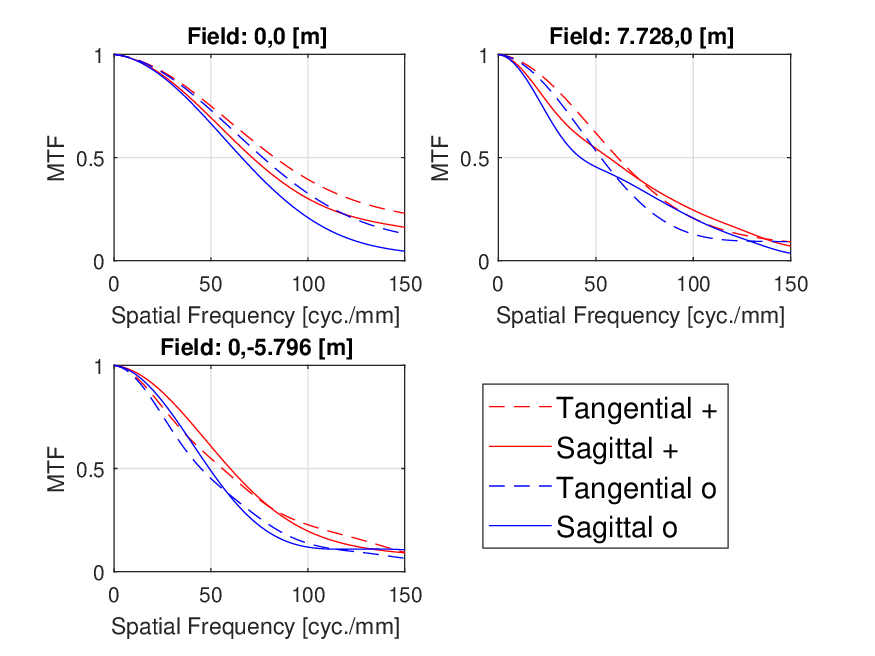}}
		\caption{MTF Comparison of `Plus'-shaped Aperture vs. Circular Aperture}
		\label{fig5}
	\end{figure}

	Figure~\ref{fig5} compares the modulation transfer function (MTF) of the two apertures. The results demonstrate that the plus-shaped aperture consistently outperforms the conventional circular aperture across all field points for both tangential and sagittal features. However, this improvement in sharpness does not necessarily imply a corresponding enhancement in the system's object detection capabilities.
	As with any deep-learning computer vision system, comparative performance analysis requires a large and diverse dataset of images containing the objects of interest (e.g., road signs, traffic lights, pedestrians, vehicles, etc.). 
	Such dataset could either be collected or synthesized. When collected, a synchronized multi-camera imaging system where each camera is equipped with a different aperture in size or shape needs to be mounted on a test-vehicle and images could be acquired during test drives in diverse locations, times and weather conditions. The acquired images would need to be annotated manually (or semi-manually) to form a ground-truth basis for training and testing the deep-learning network. 
	
	Datasets like Berkeley DeepDrive (BDD) \cite{bdd}, KITTI \cite{kitti}, and similar collections are unsuitable for this purpose due to being captured with often an unspecified optics (including the aperture shape and size). Additionally, these images typically lack per-pixel depth information, which, as will be discussed later, is essential for simulating the optical effects of the aperture. Without the depth information or a perfect knowledge of the camera optical design, it is also impossible to "deconvolve" the images to correct for optical distortions.
	
	The major leap in compute power over the past decade ignited the photo-realistic synthetic image generation for the automotive market. Synthetic images offer significant advantages over real data, including cost efficiency, scalability, and control. Synthesized images enable precise simulation of rare or hazardous scenarios, ensuring diverse datasets for robust AI training. Unlike real data, synthetic images eliminate privacy concerns, streamline labeling processes, and can be tailored to specific applications, enhancing model accuracy. 
	
	In this research the use of synthetic images was harnessed for comparative performance analysis of the various aperture shapes and their effect on the perception system object detection capabilities. 
	As shall be discussed in section~\ref{sec:psf}, emulating the optical effect of the the point spread function (PSF) is pivotal to generating photo-realistic synthetic images. 
	The PSF characterizes the response of a focused optical imaging system to a point source or object, effectively representing the system's impulse response.
	\color{black} The aperture size influences the PSF and depth of field in imaging systems. A smaller aperture increases diffraction, broadening the PSF and slightly reducing resolution but extends the depth of field, keeping more of the scene in focus. Larger apertures sharpen the PSF but narrow the depth of field\cite{AiryDisk}.\color{black}
	%The shape of the aperture and its size significantly influence the PSF by affecting the spread and sharpness of this point of light. A larger aperture allows more light and can create a smaller, sharper PSF, improving image resolution and detail. However, with smaller apertures, diffraction effects increase, causing light to spread more, which widens the PSF and reduces sharpness. This relationship means that optimal aperture size balances sharpness and light intake, as excessively small apertures blur images through diffraction, degrading the PSF and resolution.	
	%This dataset also needs to be replicated per each aperture to enable comparative performance analysis. 
	\subsection{Related Work}
	Deep learning plays a critical role in optimizing automotive camera optics and modeling the PSF. Tseng \textit{et al.} \cite{Tseng2021DeepCompoundOptics} demonstrated a framework for end-to-end camera design using differentiable optics and image processing pipelines. This approach integrates compound optics optimization with machine learning to jointly refine optical and computational components, enabling higher performance in automotive cameras. Similarly, Mosleh \textit{et al.} \cite{HWinLoop} utilized hardware-in-the-loop (HIL) methods to optimize camera image pipelines, providing real-world feedback for refining optical and algorithmic performance. Côté \textit{et al.} \cite{DiffLens} advanced compound lens design by employing differentiable models to optimize glass surfaces and materials, enhancing object detection capabilities in automotive contexts.
	
	For PSF modeling, Carney \textit{et al.} \cite{SPIE_PSF} explored deep learning to predict PSFs in complex optical systems, enabling accurate modeling for cameras under varied conditions. Herbel \textit{et al.} \cite{FastPSF} developed a fast, deep-learning-based PSF modeling approach, ensuring rapid yet precise system adjustments, vital for dynamic automotive applications. Chang and Wetzstein \cite{MonoDepth} highlighted the utility of deep optics for monocular depth estimation and 3D object detection, critical for autonomous driving. Wolf \textit{et al.} \cite{WolfWindshield} extended this by analyzing AI sensitivity to optical aberrations, emphasizing the need for robust, optimized optics in real-world environments. 
	\section{Materials and Methods}
	\label{sec:psf}
	\subsection{Emulating The Point Spread Function}
	To emulate realistic PSF values, the system parameters of the ImagingSource$^{\circledR}$ DFK37BUX252\cite{Camera} camera were used as a reference. The DFK37BUX252 is a color Industrial Camera, based on Sony 1/1.8" CMOS Pregius IMX252 sensor with a resolution of up 1536×2048 (3.1 Megapixel), with a pixel size of 3.45$\mu m$, featuring up to 119 fps with global shutter support.
	
	Given that the optical system shown in Fig.~\ref{fig1}, features a focal length of $f=16$mm, the horizontal field of view ($\textrm{FOV}_{\textrm{H}}$) of the camera equals \cite[p. 48, eq. (2.60)]{Szeliski}
	\begin{eqnarray}\label{eq:FOV_H}
		\textrm{FOV}_{\textrm{H}} = 2\cdot\tan^{-1}\left(\frac{\frac{2048}{2}\cdot3.45\cdot10^{-3}\textrm{mm}}{16\textrm{mm}}\right)=24.9^o
	\end{eqnarray}
	
	The Point Spread Function (PSF) varies based on the object's distance from the imaging plane and differs across the red, green, and blue color components. To estimate the PSF, synthetic target images were generated and processed using Zemax ray-tracing simulations, assuming object distances ranging from 10 to 100 meters in 5-meter increments (resulting in a total of 19 distances).

\begin{figure}[!ht]
		\centering
		\resizebox{3.5in}{!}{\includegraphics{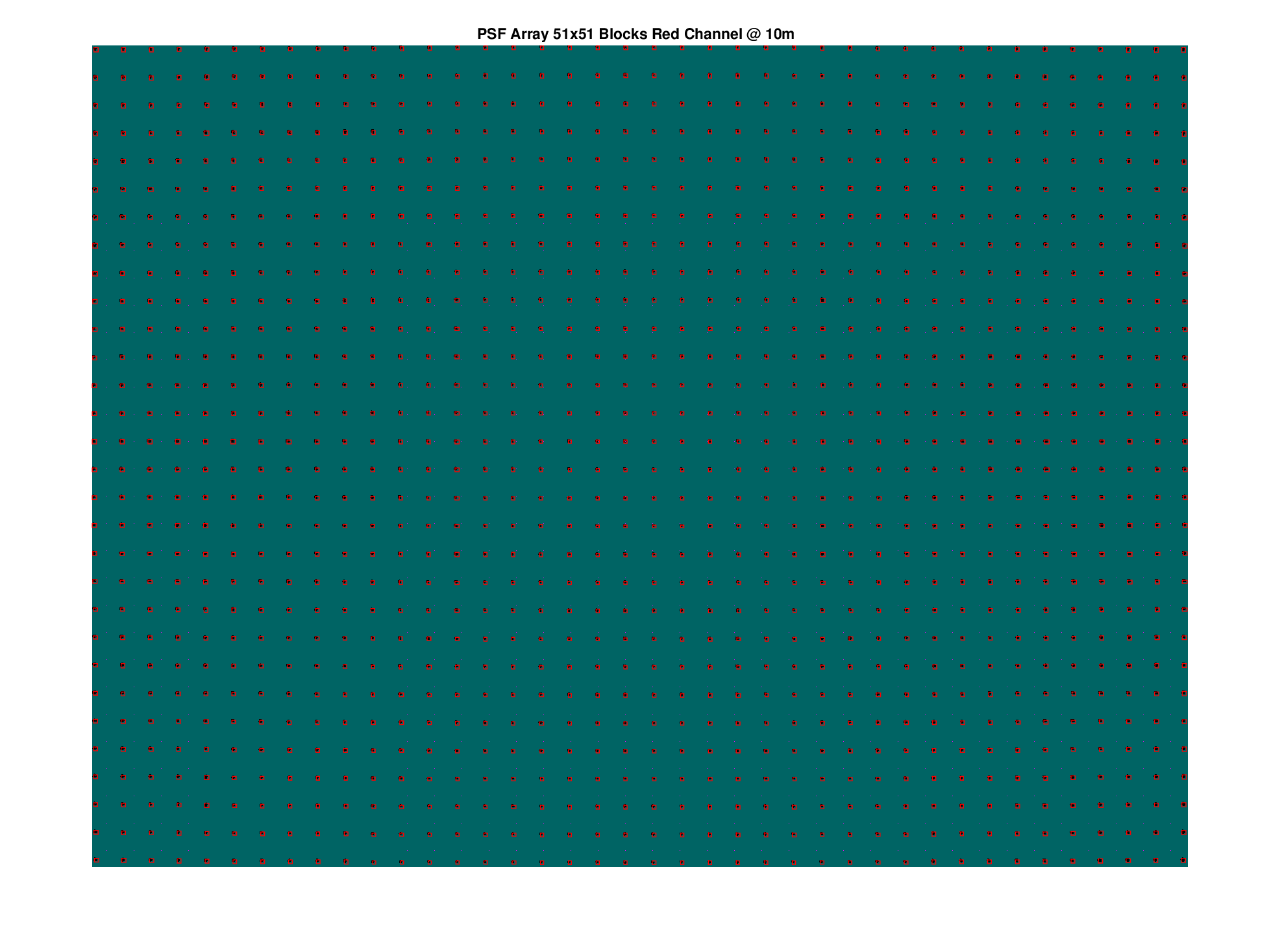}}
		\caption{PSF Array Example}
		\label{fig:fig6}
	\end{figure}

\begin{figure}[!ht]
		\centering
		%		\fbox{\rule[-.5cm]{4cm}{4cm} \rule[-.5cm]{4cm}{0cm}}
		\resizebox{3.5in}{!}{\includegraphics{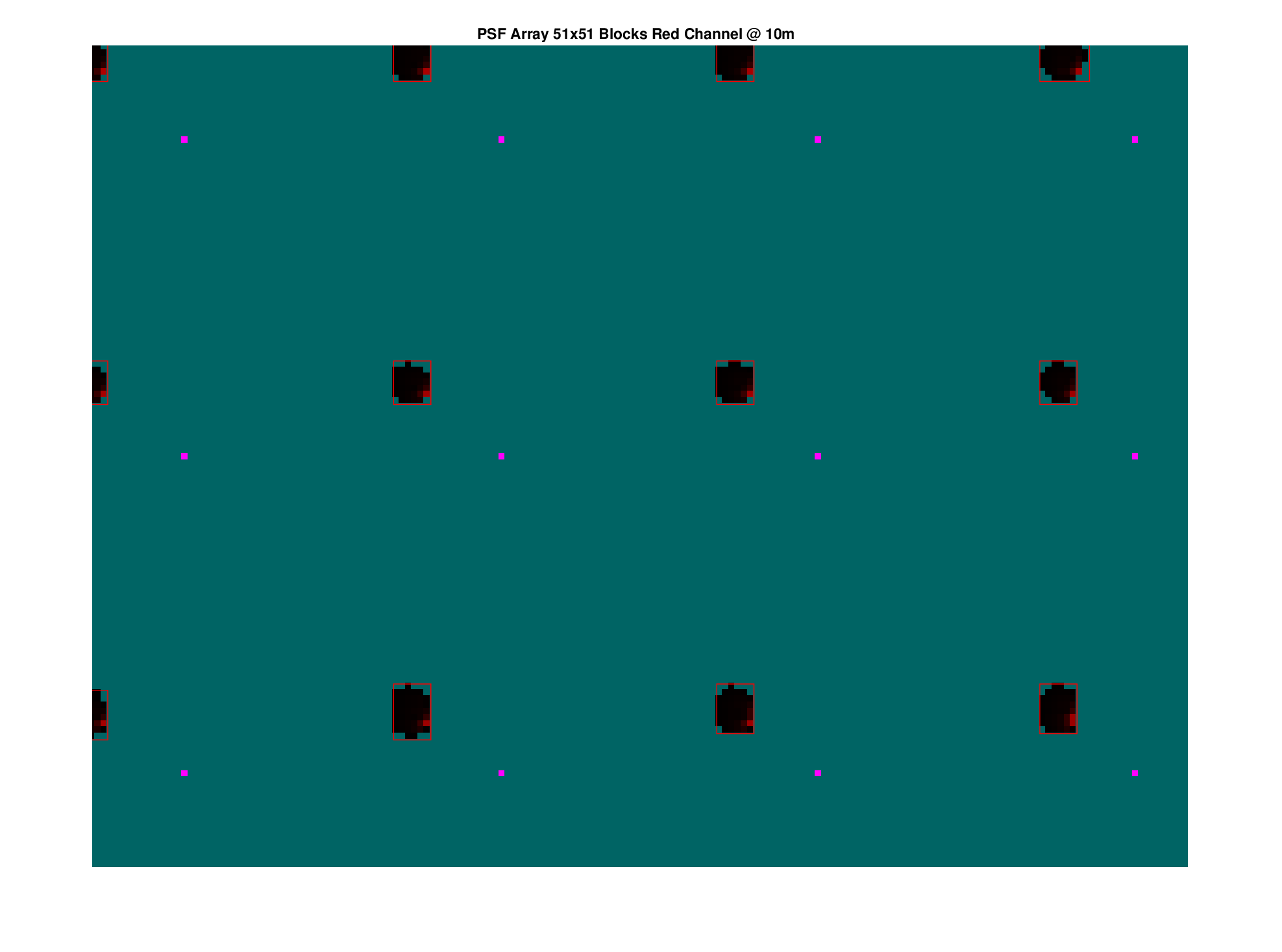}}
		\caption{PSF Array - Zoom-In Example}
		\label{fig:fig7}
	\end{figure}

\begin{figure}[!ht]
		\centering
		\resizebox{2.5in}{!}{\includegraphics{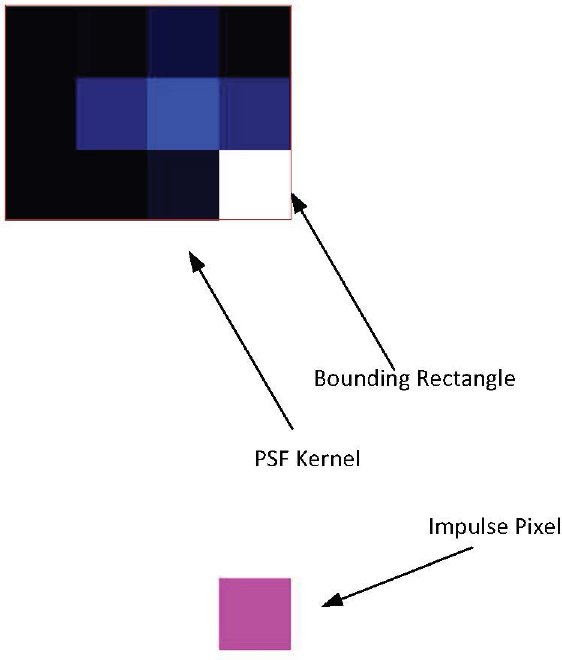}}
		\caption{PSF Array - Components}
		\label{fig:fig8}
	\end{figure}

	Each 1536$\times$2048 target image was divided into 51$\times$51 pixel blocks (with some pixel margins on the edge blocks), with all pixels set to black (hexadecimal value 0x000000), except for the center pixel of each block, which was assigned the target color component: 0xff0000 for red, 0x00ff00 for green, and 0x0000ff for blue. 
	The ray-tracing simulation produced 19 frames per color, illustrating the PSF distribution across the frame space for each distance. An example frame is shown in Fig.~\ref{fig:fig6}, with detailed views provided in Fig.~\ref{fig:fig7} and Fig.~\ref{fig:fig8}. In these figures, the magenta-colored pixels indicate the original impulse location, while the PSF is represented by a pixel blob enclosed within a rectangular box, displaying varying intensity levels. This blob signifies the average PSF within the block. The spatial displacement of the blob relative to the original impulse results from refraction effects caused by the vehicle windshield.

	\subsection{Converting ``Pinhole''-Images to PSF-Filtered }
\begin{figure}[!ht]
		\centering
		%		\fbox{\rule[-.5cm]{4cm}{4cm} \rule[-.5cm]{4cm}{0cm}}
		\resizebox{3.5in}{!}{\includegraphics{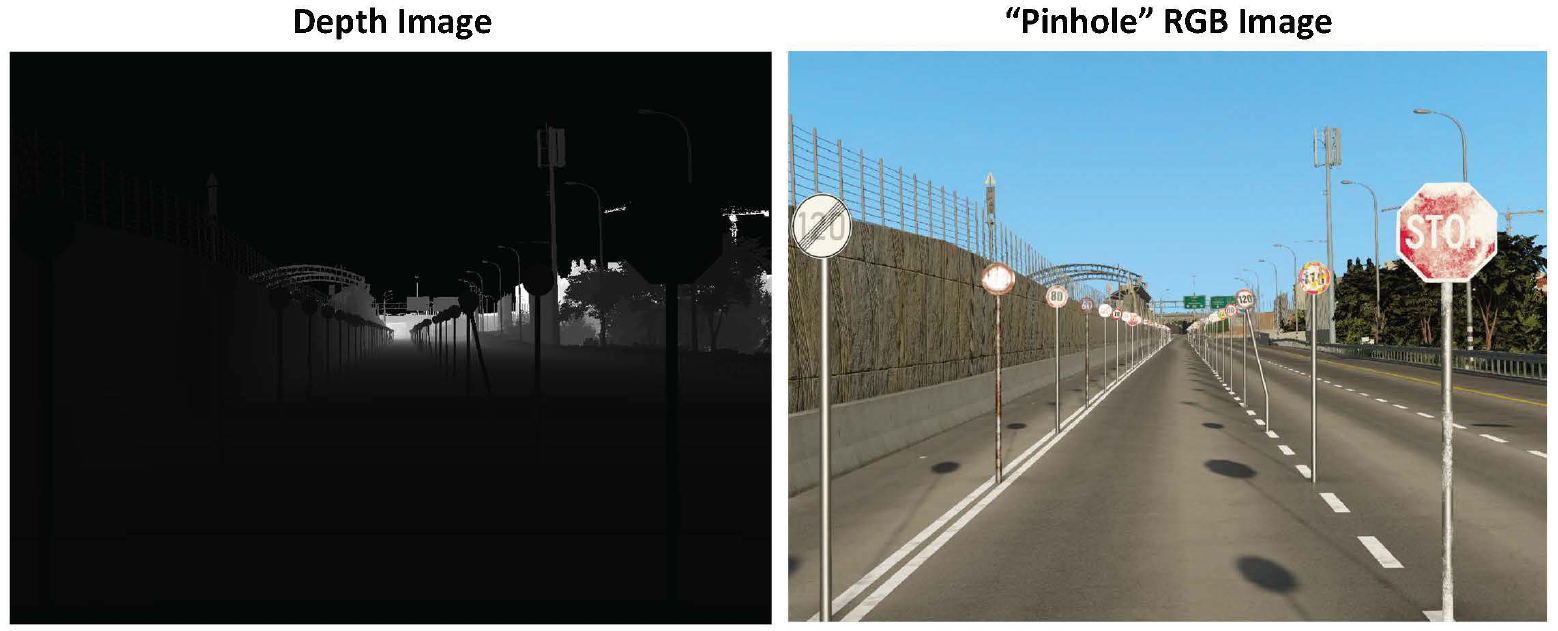}}
		\caption{Pinhole Image and its Associate Depth Image}
		\label{fig:fig13}
	\end{figure}

	The synthetic images originally generated for the dataset were free of optical aberrations. In addition to being fully annotated, each image was accompanied by a per-pixel 16-bit depth map that specified the depth of each pixel in meters. Fig.~\ref{fig:fig13} illustrates an example of a pinhole RGB image alongside its corresponding depth map.
	
\begin{figure}[!ht]
		\centering
		%		\fbox{\rule[-.5cm]{4cm}{4cm} \rule[-.5cm]{4cm}{0cm}}
		\resizebox{3.5in}{!}{\includegraphics{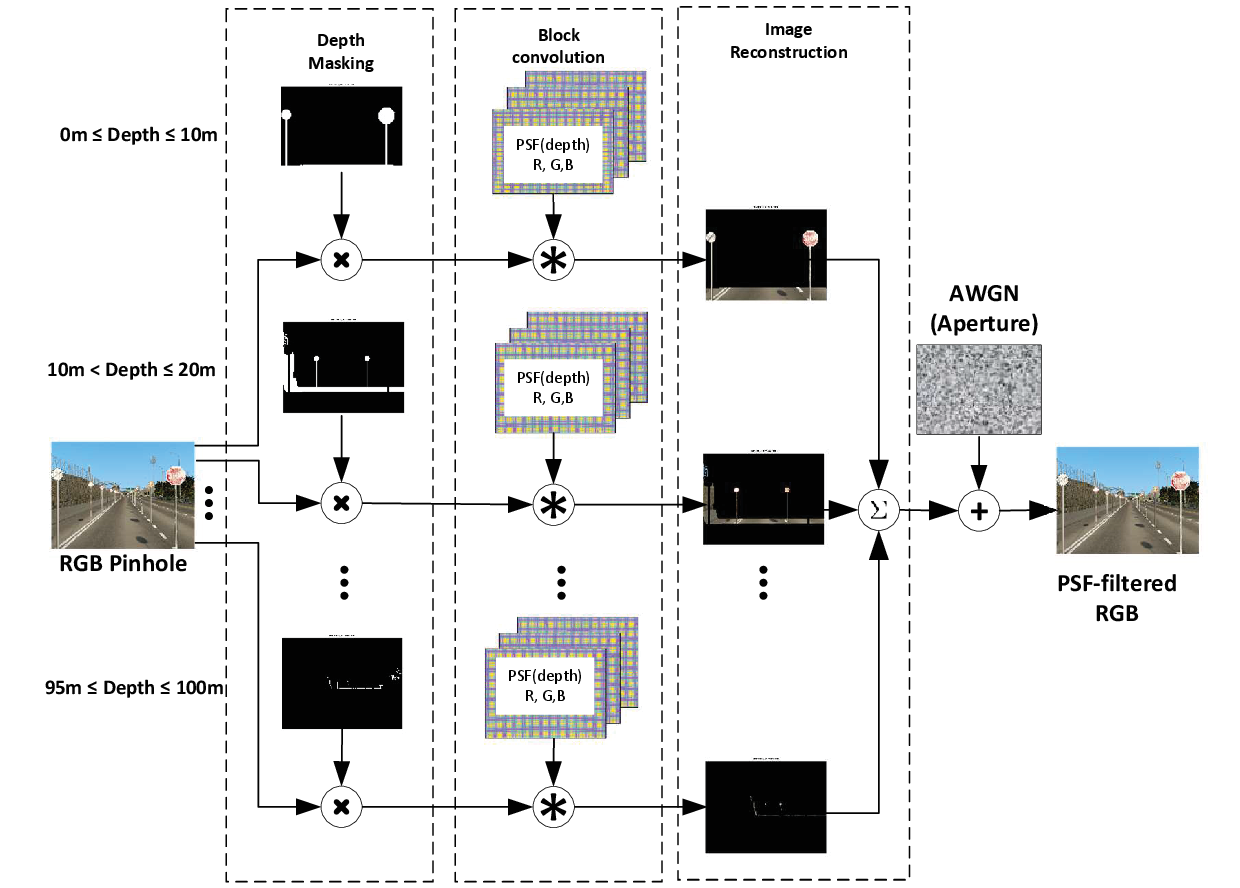}}
		\caption{PSF-Filtering Pipeline}
		\label{fig:fig4}
	\end{figure}

\begin{figure}[!ht]
		\centering
		%		\fbox{\rule[-.5cm]{4cm}{4cm} \rule[-.5cm]{4cm}{0cm}}
		\resizebox{2.5in}{!}{\includegraphics{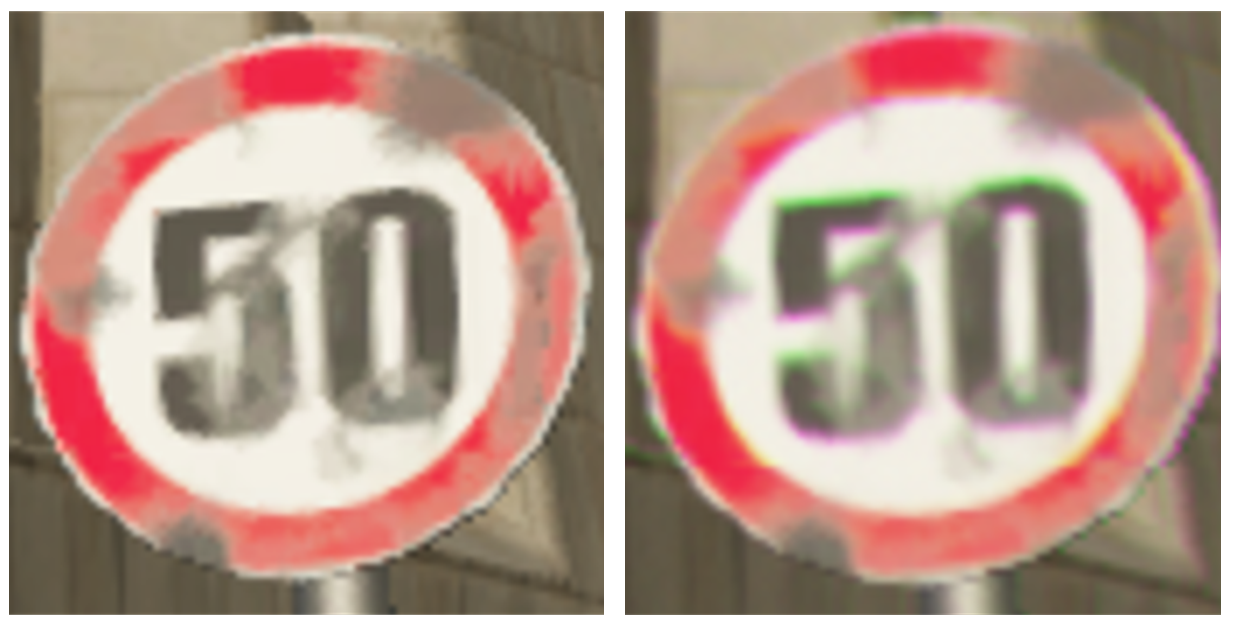}}
		\caption{Pinhole (left) vs. PSF-Filtered Image (right)}
		\label{fig:fig9}
	\end{figure}

\begin{figure}[!ht]
		\centering
		%		\fbox{\rule[-.5cm]{4cm}{4cm} \rule[-.5cm]{4cm}{0cm}}
		\resizebox{3.5in}{!}{\includegraphics{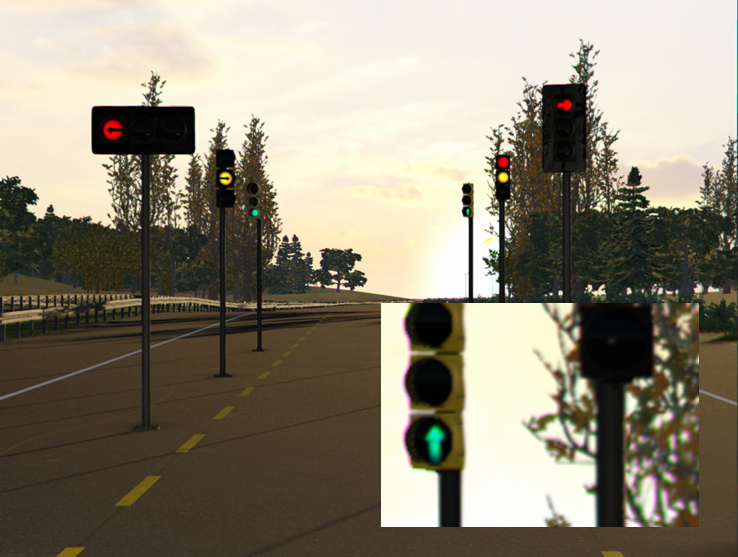}}
		\caption{Pinhole vs. PSF-Filtered Image in Traffic Light Objects}
		\label{fig:fig10}
	\end{figure}

	The depth buffer was utilized to exclude pixels outside the relevant range for depth-dependent PSF filtering. The remaining pixels were then filtered color by color using the PSF ``blob'' filters. This process was repeated across all 19 depth ranges, and the filtered images were subsequently combined to reconstruct a single image. Before producing the final output, Gaussian noise was added to account for differences in aperture cross-sections, as discussed further in section~\ref{sec:psf_awgn}. This workflow is depicted in Fig.~\ref{fig:fig4}, and examples of filtered images, showcasing chromatic aberrations at the edges caused by the aperture shape, are presented in Fig.~\ref{fig:fig9}-\ref{fig:fig10}.
	%\clearpage
	\subsection{PSF Relative Noise Balancing}
	\label{sec:psf_awgn}
	\color{black}
	Apart from the diffraction effects introduced by the PSF, a smaller aperture cross-section reduces the amount of light reaching the imaging sensor, resulting in darker images. To counteract this, one can assume the images maintain consistent intensity by adjusting the gain based on the aperture cross-section. By applying PSF filters of varying shapes and compensating with appropriate gain levels, the intensity of the images can remain uniform. However, this approach has a trade-off: increasing the gain to offset reduced light also amplifies additive noise, potentially degrading the image quality. 
	\color{black}

	\noindent \color{black}The required gain to be applied for maintaining the same level of intensity across different apertures depends on \color{black} the f-number of the lens (commonly denoted as $f/\#$), which is defined as 
	\begin{eqnarray}
		f/\#=\frac{f}{D}
	\end{eqnarray}
	where $D$ is the effective aperture diameter and $f$ is its focal length (both specified in [mm]). 
	
	Table~\ref{tab:apetures_params} summarizes the parameters of the four apertures analyzed in this research. The circular aperture featuring a diameter of 9mm, serves as the reference, with a cross-sectional area of $4.5^2\cdot\pi=$63.6mm$^2$.  Recall that the focal length is 16mm, the nominal f-number of the circular aperture is $f/\#^{\textrm{circular}}=16\textrm{mm}/9\textrm{mm}\approx1.8$. The f-numbers of the remaining apertures are calculated relative to the circular aperture.
	\begin{eqnarray}
		f/\#^{\textrm{aperture}}\approx f/\#^{\textrm{circular}}\sqrt{\frac{\textrm{Area}_{\textrm{circular}}}{\textrm{Area}_{\textrm{aperture}}}}
	\end{eqnarray}
	Similarly, the gain required to be applied to images processed with non-circular apertures to maintain the same level of intensity is calculated as
	\begin{eqnarray}
		\textrm{Gain Factor [dB]}= 20\log_{10}\left(\frac{f/\#^{\textrm{aperture}}}{f/\#^{\textrm{circular}}}\right)
	\end{eqnarray}
	\begin{table}[!ht]
		\centering
		\scriptsize
		\begin{tabular}{llcccc}
			\midrule
			\textbf{No.} &\textbf{Aperture Shape} &	\textbf{Area [mm$^2$]} &	\textbf{Area factor}	& {$\mathbf{f/\#}$}  &	\textbf{Gain Factor [dB]}\\
			\midrule
			1 &circular            &	63.6          &	1.00        &	1.8	  & 0.0 \\
			2 & plus               &	35.6          &	0.56        &	2.4	  & 5.1\\
			3 & vertical slit	   &    17.6          &	0.28	    &   3.4	  & 11.2\\
			4 & horizontal slit	   &    17.6	      & 0.28	    &   3.4	  & 11.2\\
			\bottomrule
		\end{tabular}
		\caption{Apertures Parameters}
		\label{tab:apetures_params}
	\end{table}
	
\begin{figure}[!ht]
		\centering
		%		\fbox{\rule[-.5cm]{4cm}{4cm} \rule[-.5cm]{4cm}{0cm}}
		%\resizebox{5.5in}{!}{\includegraphics{noise_vs_gain.eps}}
		\resizebox{3.5in}{!}{\includegraphics{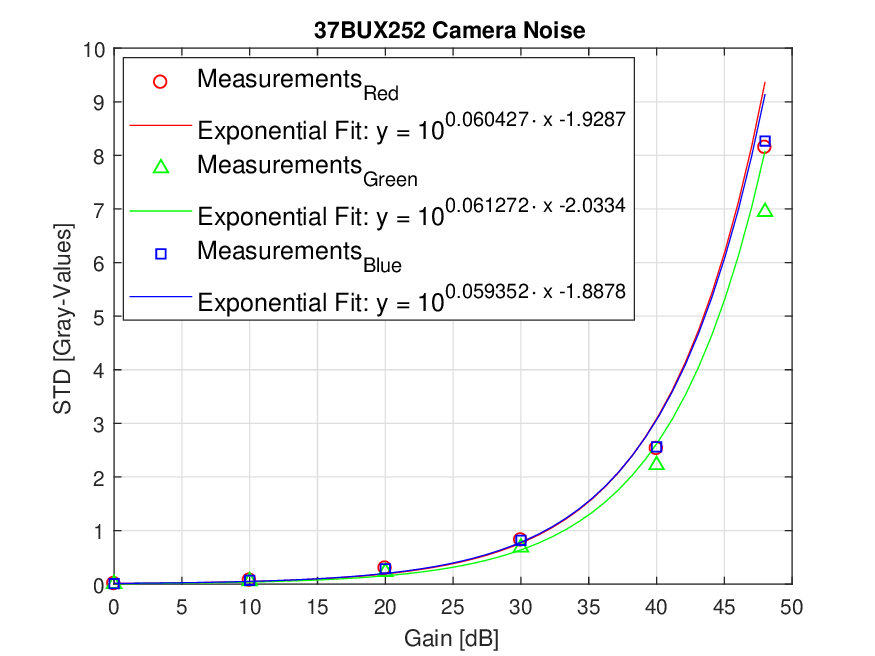}}
		\caption{Measured Camera Noise vs. Camera Gain}
		\label{fig:fig25}
	\end{figure}

	Again, the DFK37BUX252 camera was used as a reference to simulate realistic noise.  
	Figure~\ref{fig:fig25} illustrates the relationship between camera gain and the noise standard deviation (STD) expressed in 8-bit gray values (0–255) per color component. The noise STD was calculated across multiple frames captured at 60 frames per second (fps) with the camera lens obscured, while gain levels ranging from 0 to 48 dB were applied. To interpolate the gain factors listed in Table~\ref{tab:apetures_params}, an exponential curve was fitted to the measured STD values.

	%\clearpage

	\subsection{Dataset Generation}
	To circumvent the need to collect and manually annotate the images, a dataset of synthetic images containing the target object classes was created. The original ``pinhole'' dataset (i.e., images without any optical distortion) comprised approximately 41,000 images. This dataset included images of traffic lights, traffic signs, and speed signs in equal proportions. The images were rendered using the Cognata traffic sign dataset \cite{CognataTS} and its counterpart, the traffic light dataset \cite{CognataTL}, both adhering to the European format. The target objects were depicted in various geographic locations, under diverse lighting conditions (e.g., dawn, morning, noon, afternoon, sunset, and night), and across different weather scenarios (e.g., clear skies, partly cloudy, overcast, foggy, etc.). The objects appeared as columns with fixed spacing along the roadside, at distances ranging from 10–20 meters to over 100 meters.
	Each image in the dataset was fully auto-annotated, segmented, and accompanied by per-pixel depth information, as illustrated in Fig.~\ref{fig:fig13}.
	
\begin{figure}[!ht]
		\centering
		%		\fbox{\rule[-.5cm]{4cm}{4cm} \rule[-.5cm]{4cm}{0cm}}
		\resizebox{3.5in}{!}{\includegraphics{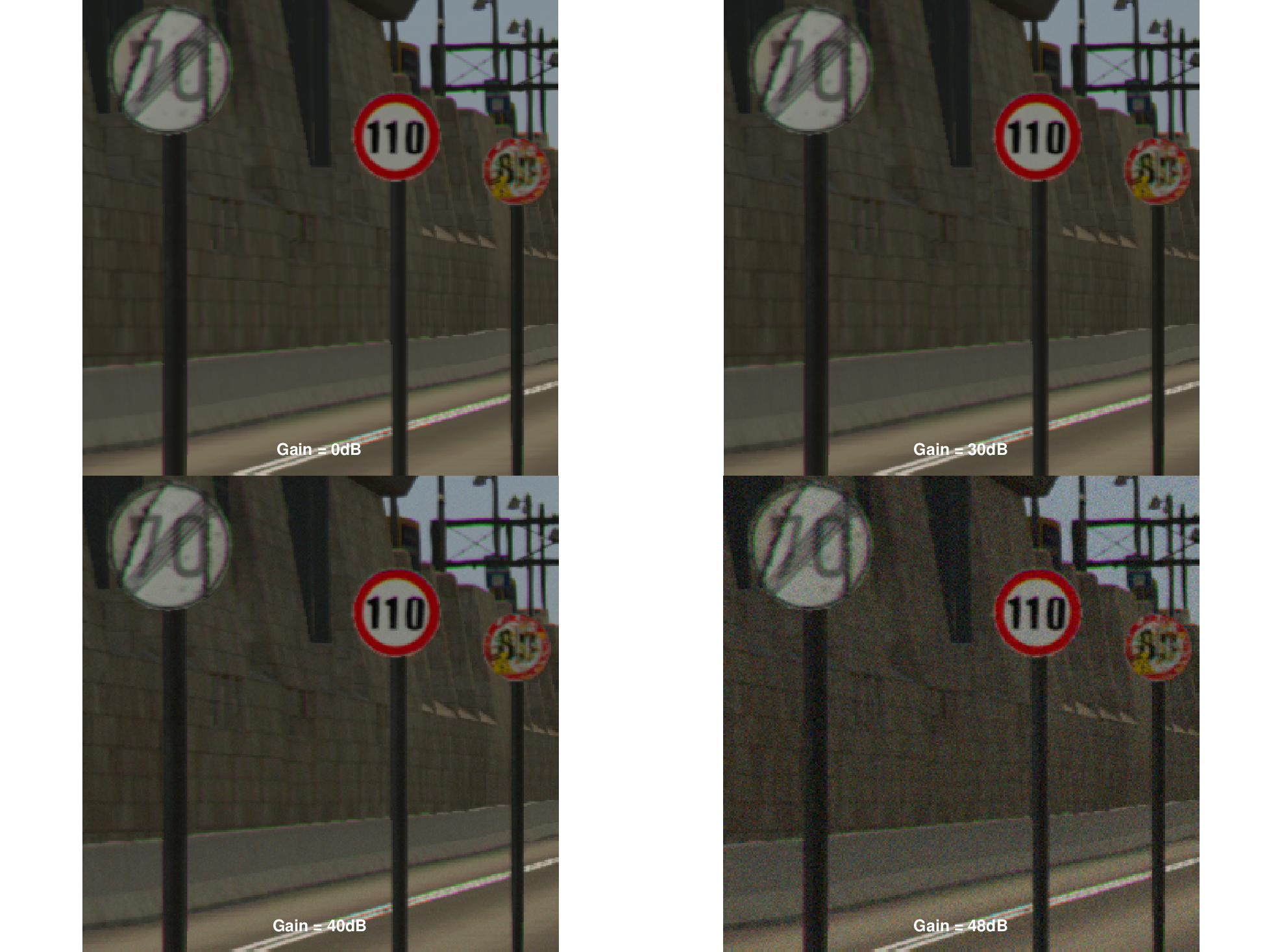}}
		\caption{Speed Sign Image Example with Four Levels on Camera Gain}
		\label{fig:fig28}
	\end{figure}

\begin{figure}[!ht]
		\centering
		%		\fbox{\rule[-.5cm]{4cm}{4cm} \rule[-.5cm]{4cm}{0cm}}
		\resizebox{3.5in}{!}{\includegraphics{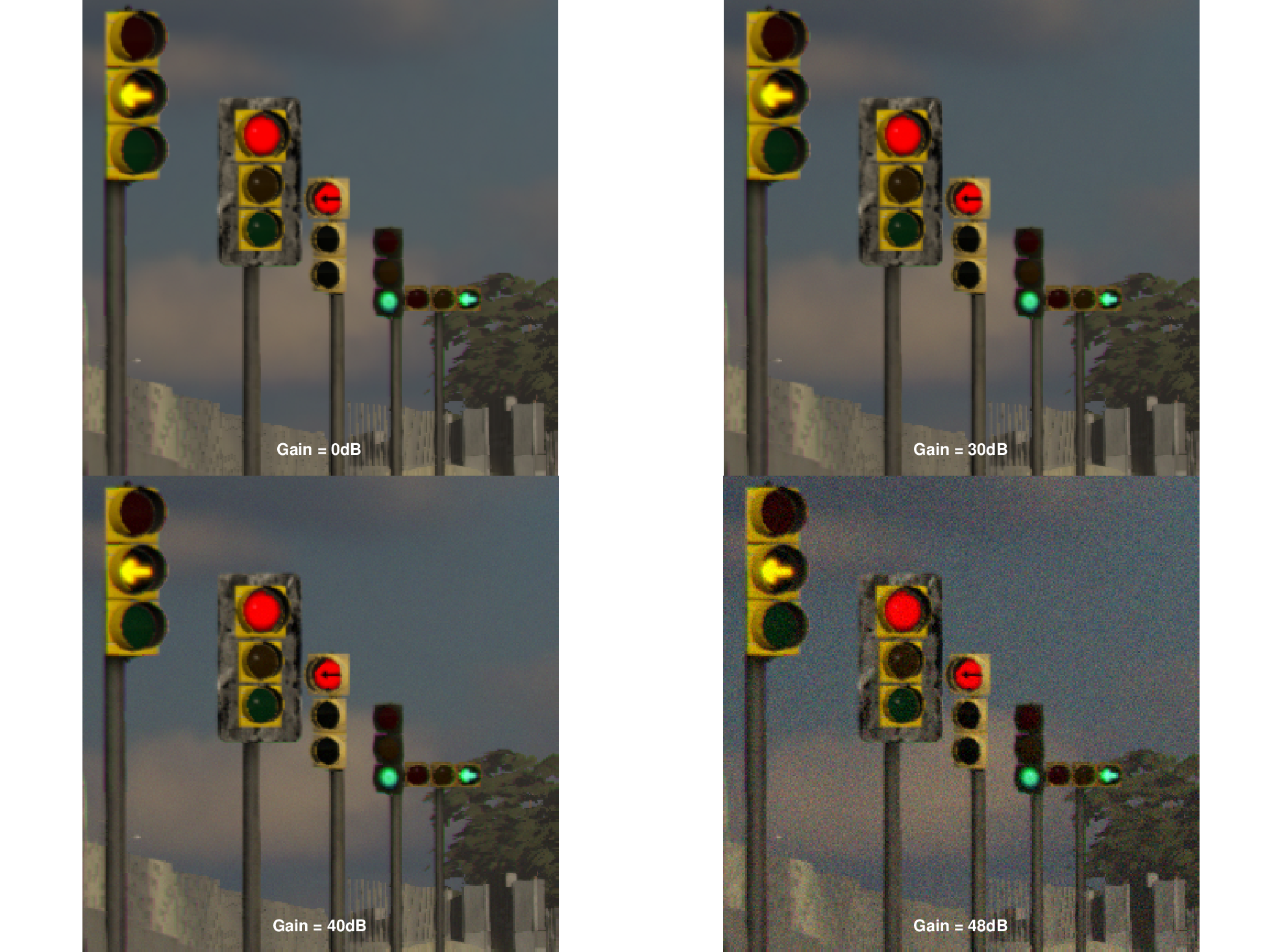}}
		\caption{Traffic Light Image Example with Four Levels on Camera Gain}
		\label{fig:fig29}
	\end{figure}

	The ``pinhole'' dataset was expanded into 16 replicas. Initially, it was processed with PSF filtering, as described in Section~\ref{sec:psf}, using the emulated PSFs of the four apertures listed in Table~\ref{tab:apetures_params}. Furthermore, three replicas for each aperture were augmented with zero-mean additive white Gaussian noise (AWGN) at camera gain levels of 30 dB, 40 dB, and 48 dB. The AWGN standard deviation was determined based on the measurements shown in Fig.~\ref{fig:fig25}. Together with a noise-free replica (corresponding to 0 dB gain), these augmentations resulted in a total of 16 replicas derived from the original dataset.
	Figures~\ref{fig:fig28}-\ref{fig:fig29} depict examples of speed signs and traffic lights images under the various camera gain/additive noise conditions. 
	\subsection{Network Configuration}
\begin{figure}[!ht]
		\centering
		%		\fbox{\rule[-.5cm]{4cm}{4cm} \rule[-.5cm]{4cm}{0cm}}
		\resizebox{2.5in}{!}{\includegraphics{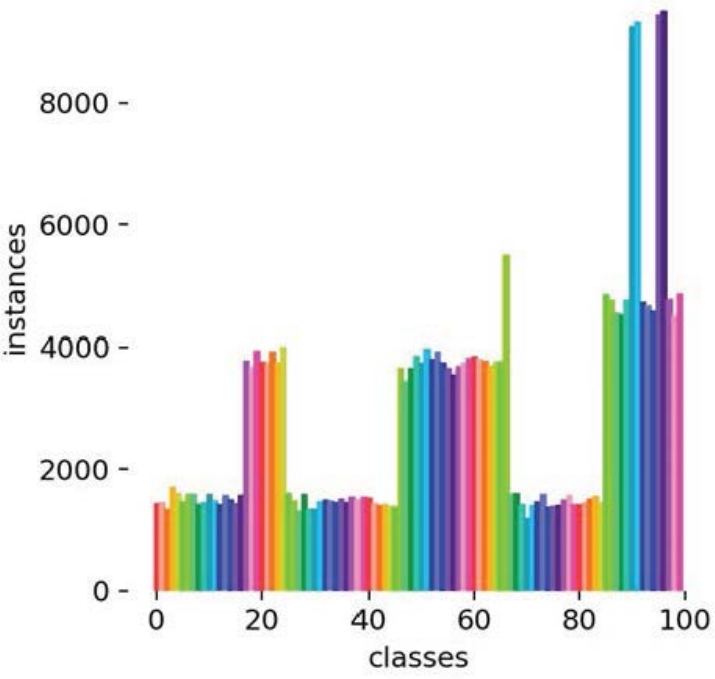}}
		\caption{Classes Distribution}
		\label{fig:fig30}
	\end{figure}

	In this research, the Ultralytics YOLOv8 deep neural network (DNN)\cite{YOLOv8, YOLOorig}  was trained for the object detection tasks. The ``Nano'' version of YOLOv8 (YOLOv8n), featuring the smallest memory footprint was used. 
	\color{black}
	The DNN was configured to detect and classify 100 unique classes of both traffic-signs, traffic lights and speed signs. Fig.~\ref{fig:fig30} describes the distribution of instances per class within the dataset images. The traffic signs include 57 classes, (indexed in Fig.~\ref{fig:fig30} from 0-16, 25-45, and 66-84), speed-signs include 28 classes (indexed from 17-24 and 56-65). Finally, the traffic lights include 15 classes (indexed from 85-99). 
	\color{black}

	\section{Results}

	%\subsection{Numerical Analysis}
	The DNN object detection precision performance was cross-validated using K-fold methodology\cite{Kfold} with $K=5$.
	A unique YOLOv8 DNN was trained for each combination of camera aperture, camera gain, and K-fold (for 4 apertures, 4 camera gain levels and 5-fold cross validation a total of 80 YOLOv8 networks were trained and tested). Each fold of 41,000 images was split into 80\% of the images (randomly selected) that were used for training, 10\% for validation and 10\% for testing. 
	
	The analysis was done separately per class group (speed signs, traffic signs and traffic lights), and bounding-box size according to the sizes defined by Microsoft COCO\cite{coco}, (with the addition of the ``tiny''-sized bounding box (of 23$\times$23 pixels), excluded from the original COCO code). To provide context for the bounding box sizes, Table~\ref{tab:bbox_range} specifies the corresponding distances for each object type and bounding box size, with the range calculation detailed in Appendix~\ref{sec:depth_calc}.
	\begin{table*}[t]
		\small
		\centering
		\begin{tabular}{lllll}
			\hline
			\textbf{bbox type}    & \makecell{\textbf{bbox size}\\ $[$\textbf{pixels}$]$} & \makecell{\textbf{Distance to}\\ \textbf{Speed Sign}} &  \makecell{\textbf{Distance to}\\ \textbf{Traffic Light}} &  \makecell{\textbf{Distance to}\\ \textbf{Traffic Sign}} \\
			\hline
			tiny     & bbox $\leq$ 23$\times$ 23                   &Dist. > 51m  & Dist. > 63m & Dist. > 129m\\
			small    & 23 $\times$ 23 < bbox $\leq$ 32$\times$ 32​     & 38m < Dist. <  51m
			& 46m < Dist. <  63m & 93m < Dist. <  129m​     \\
			medium   & 32$\times$ 32 < bbox $\leq$ 96$\times$ 96​      & 14m <  Dist. < 38m & 16m <  Dist. < 46m & 31m <  Dist. < 93m      \\
			large    & bbox > 96$\times$96​                        & Dist. < 14m & Dist. < 16m
			&  Dist. < 31m    \\
			\hline
		\end{tabular}
		\caption{Bounding Box Sizes \& Corresponding Object Distance}
		\label{tab:bbox_range}
	\end{table*}

	The metric used for the comparative performance analysis was the mean average precision (mAP)
	\begin{eqnarray}
		\textrm{mAP}=\frac{1}{|\textrm{classes}|}\sum_{c\in \textrm{classes}}\frac{|TP_c|}{|TP_c| + |FP_c|}
	\end{eqnarray}
	where $|TP_c|$ denotes ``true-positive'' counts for the $c$th class and $|FP_c|$ denotes ``false positive'' counts for the $c$th class. A \textit{true-positive} event is counted if two conditions are met: the confidence score of the predicted bounding box is greater than the confidence threshold, and the intersection-over-union (IoU) between the predicted bounding box and the ground-truth bounding box is greater than the IoU threshold. A detection would be considered as \textit{false-positive} event if either: the model detects an object with a high confidence score, but it is not present (no ground truth), then the IoU would be equal to zero.
	Alternatively, a false-positive event will be counted if the IoU is less than the IoU threshold, or if the proposed bounding box aligns with the ground truth, but the class label of the proposed box is incorrect. 
	
\begin{figure}[!ht]
		\centering
		%		\fbox{\rule[-.5cm]{4cm}{4cm} \rule[-.5cm]{4cm}{0cm}}
		\resizebox{3.5in}{!}{\includegraphics{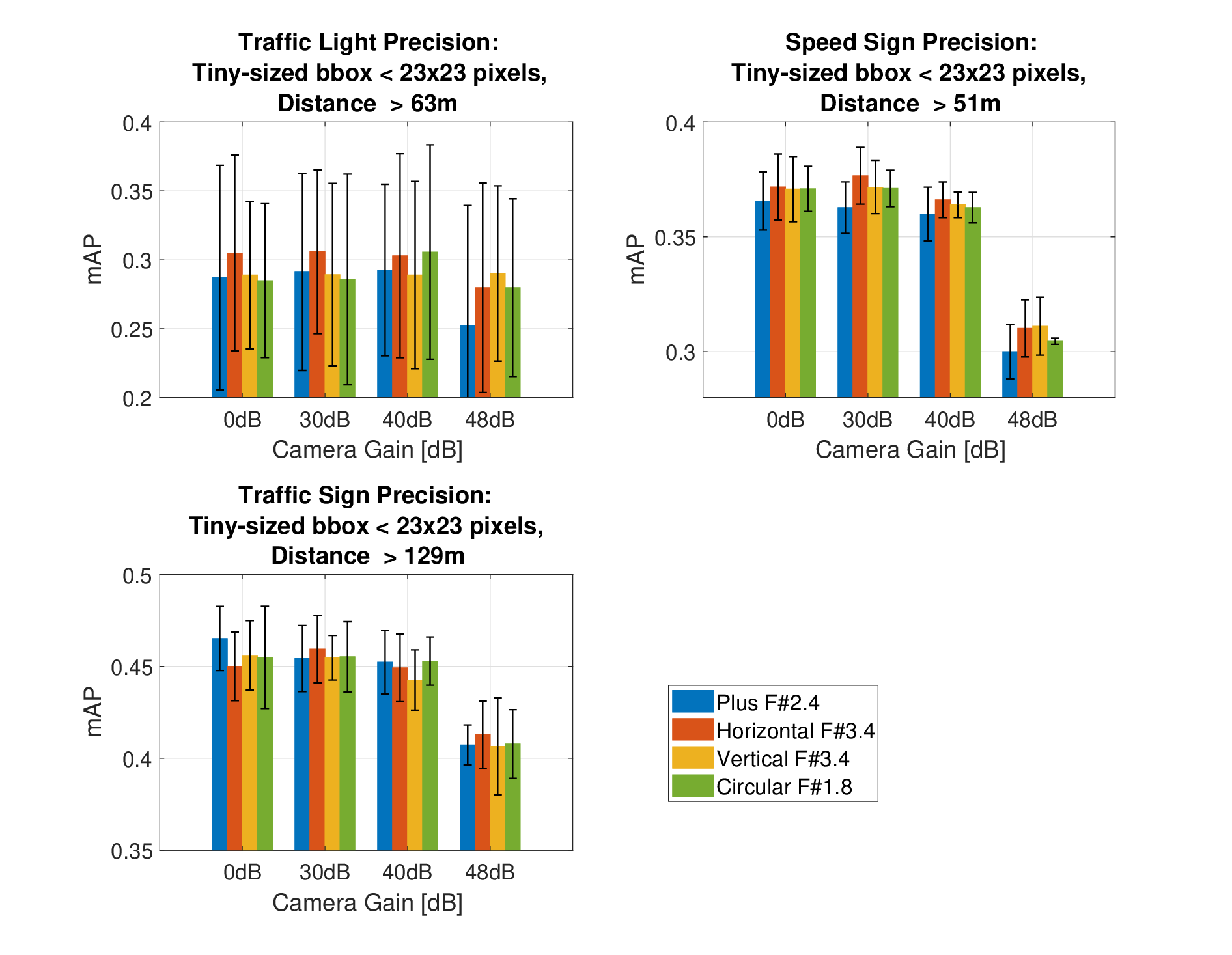}}
		\caption{mAP for Tiny-Sized Bounding Boxes}
		\label{fig:fig11}
	\end{figure}

\begin{figure}[!ht]
		\centering
		%		\fbox{\rule[-.5cm]{4cm}{4cm} \rule[-.5cm]{4cm}{0cm}}
		\resizebox{3.5in}{!}{\includegraphics{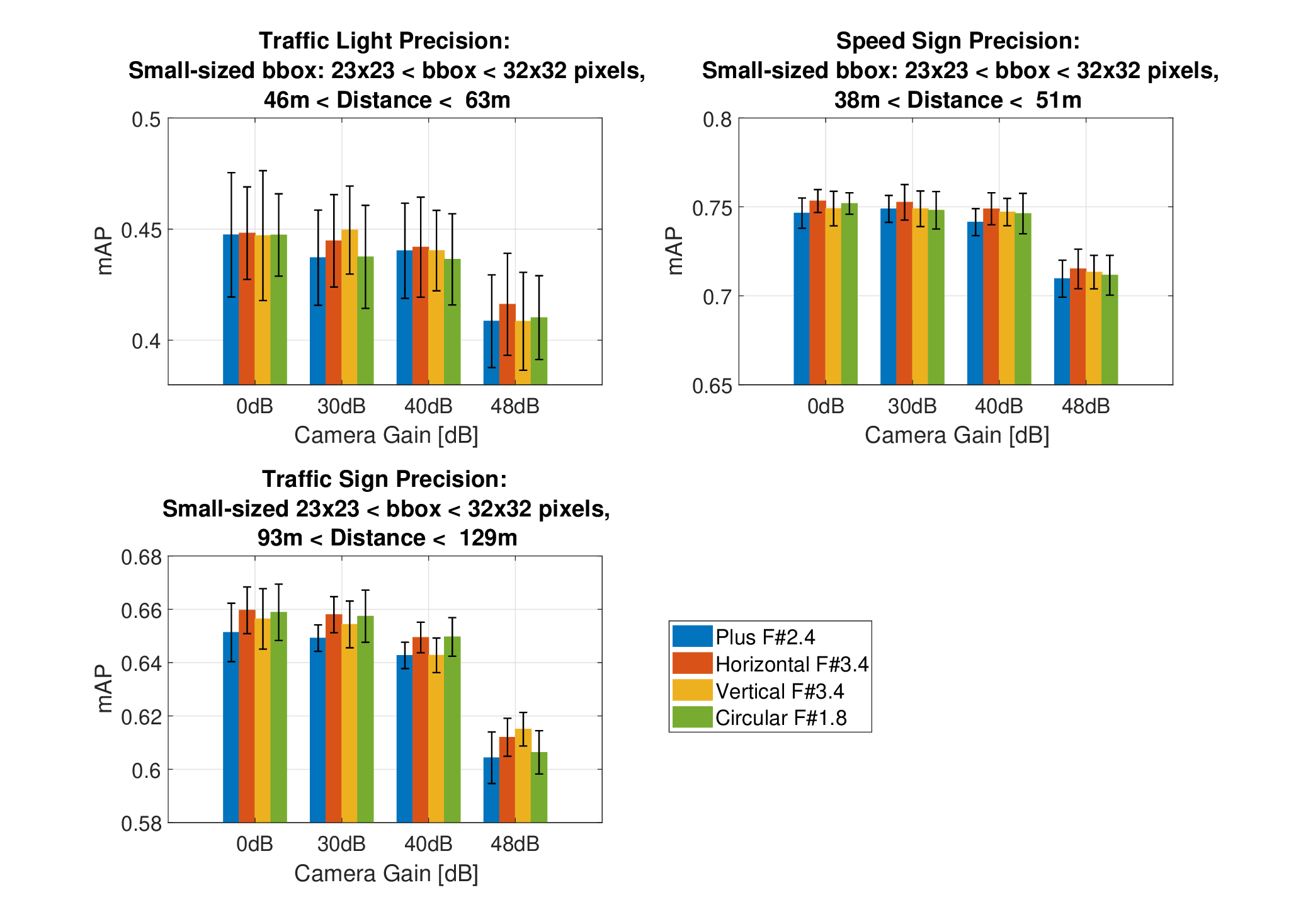}}
		\caption{mAP for Small-Sized Bounding Boxes}
		\label{fig:fig12}
	\end{figure}

\begin{figure}[!ht]
		\centering
		%		\fbox{\rule[-.5cm]{4cm}{4cm} \rule[-.5cm]{4cm}{0cm}}
		\resizebox{3.5in}{!}{\includegraphics{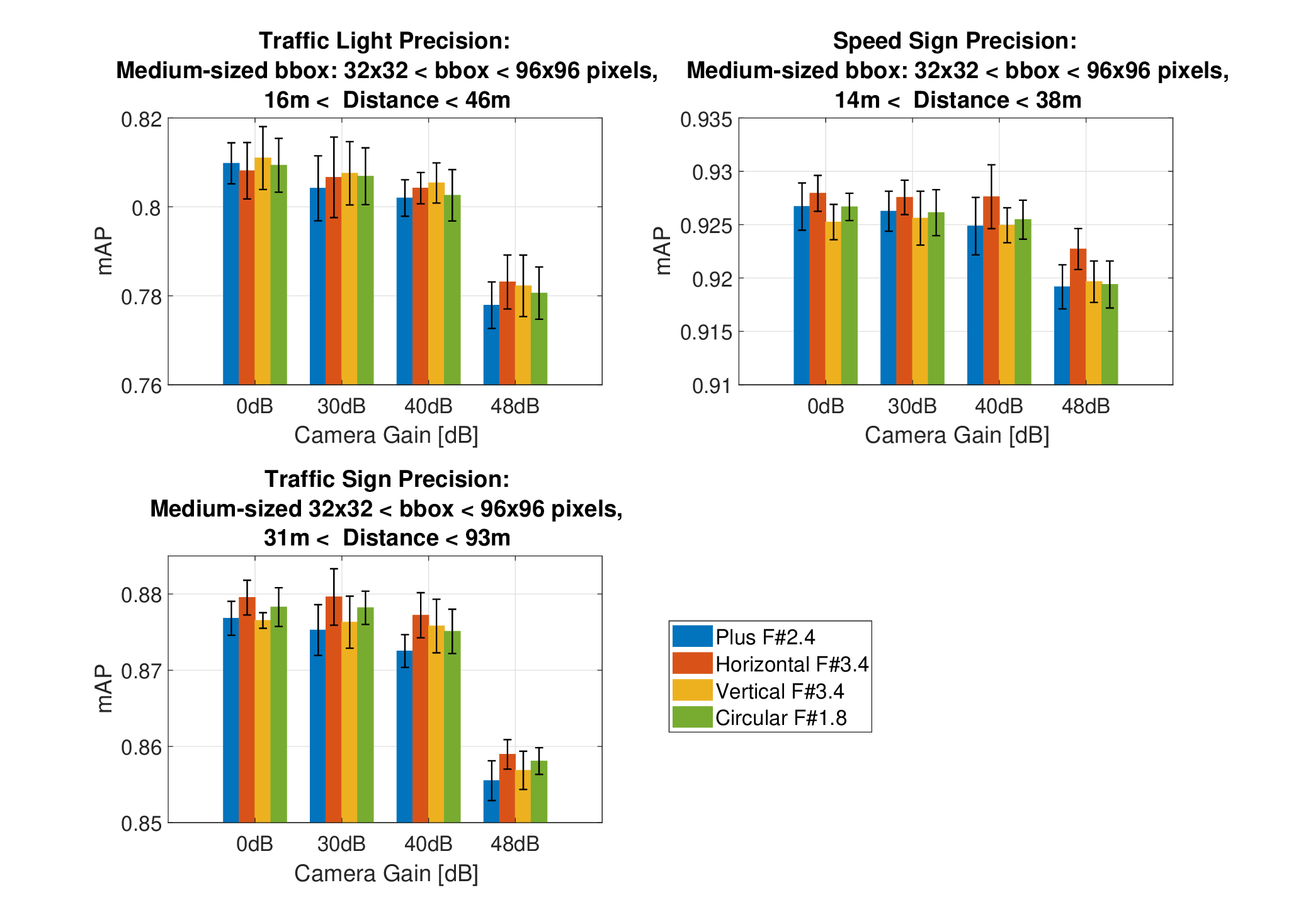}}
		\caption{mAP for Medium-Sized Bounding Boxes}
		\label{fig:fig14}
	\end{figure}

\begin{figure}[!ht]
		\centering
		%		\fbox{\rule[-.5cm]{4cm}{4cm} \rule[-.5cm]{4cm}{0cm}}
		\resizebox{3.5in}{!}{\includegraphics{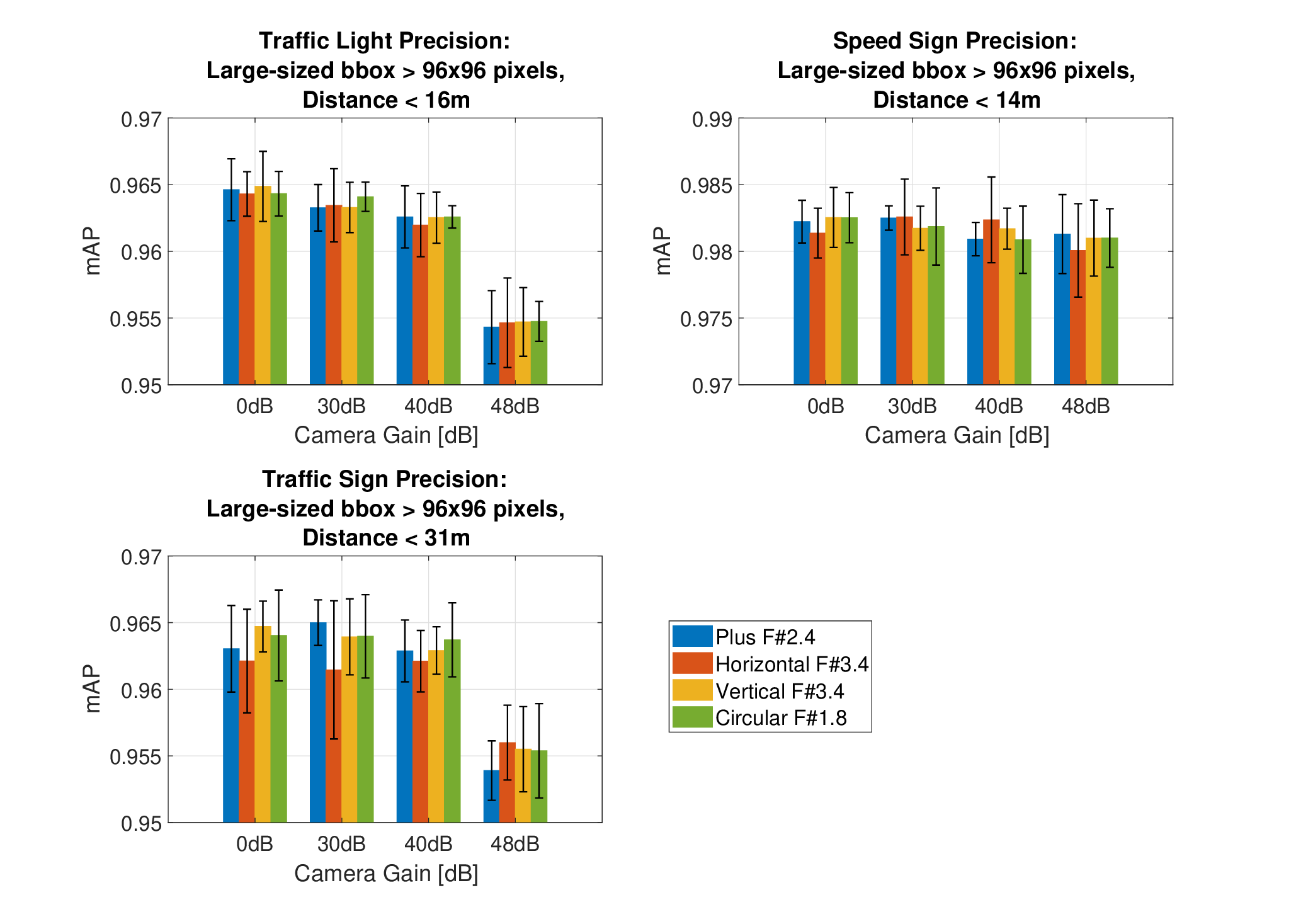}}
		\caption{mAP for Large-Sized Bounding Boxes}
		\label{fig:fig15}
	\end{figure}

\begin{figure}[!ht]
		\centering
		%		\fbox{\rule[-.5cm]{4cm}{4cm} \rule[-.5cm]{4cm}{0cm}}
		\resizebox{3.5in}{!}{\includegraphics{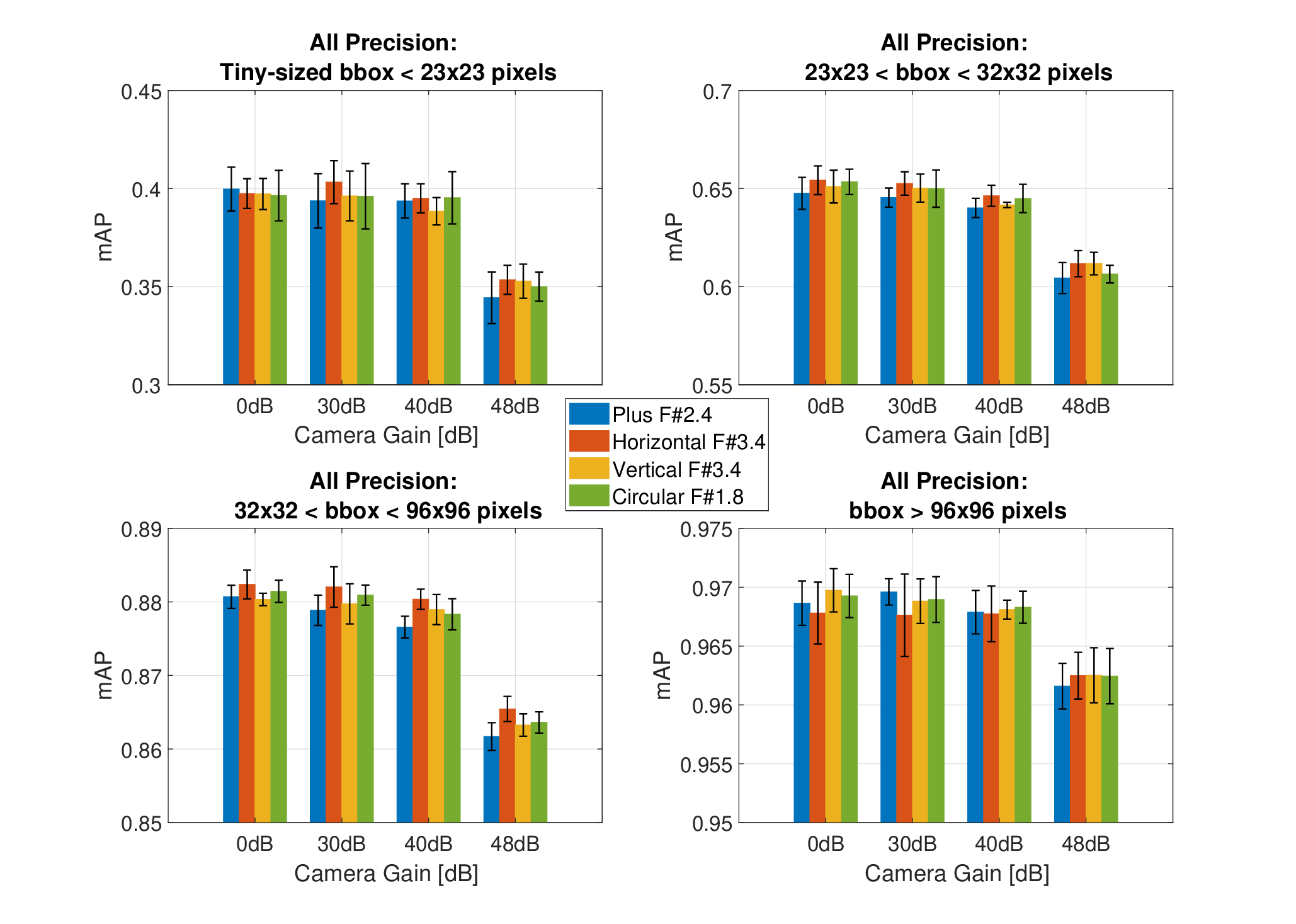}}
		\caption{mAP for All Classes Combined}
		\label{fig:fig16}
	\end{figure}

	The bars height in Figs.~\ref{fig:fig11}-\ref{fig:fig16} correspond to the  K-fold weighted-mean mAP calculated for IoU threshold of the range 0.50-0.95 in steps of 0.05 (as commonly used by COCO), for the class groups of traffic lights, traffic signs and speed signs. The mAP results correspond to bounding box (bbox) sizes as specified in Table~\ref{tab:bbox_range}. The bar heights were calculated using  
	\begin{eqnarray}
		\bar{\mu}^*=\sum_{k=1}^K w_k\cdot{\mu}_k
	\end{eqnarray}
	where ${\mu}_k$ denotes the mAP of the $k$th fold calculated using \verb|cocoeval.py|\footnote{\url{https://github.com/cocodataset/cocoapi/blob/master/PythonAPI/pycocotools/cocoeval.py}} for IoU threshold of the range 0.50-0.95, and $w_k$ denotes the weight of the $k$th fold, which is the fold count of the bounding boxes corresponding to the relevant size and to the relevant class category (traffic sign, traffic light or speed sign).
	The error-bars represent the weighted standard deviation (STD) of K-fold, which is given by \cite{Dataplot}
	\begin{eqnarray}
		\bar{\sigma} = \sqrt{
			\frac{
				\sum_{k=1}^K w_k \left( {\mu}_k - \bar{\mu}^* \right)^2
			}{
				\frac{K-1}{K} \sum_{k=1}^K w_k
			}
		}
	\end{eqnarray}
	Analyzing figs.~\ref{fig:fig11}-\ref{fig:fig15}, it becomes evident that the STD for tiny and small bounding boxes is notably higher than that for medium and large bounding boxes, with the tiny traffic light cases being particularly pronounced. This discrepancy arises from the image synthesis process used by the Cognata simulator, where objects are arranged in columns along the roadside. As a result, the dataset contains significantly fewer tiny bbox examples compared to other sizes: approximately 1\% are tiny bounding boxes, ~19\% are small bounding boxes, ~30\% are large bboxes, and the remaining ~50\% are medium-sized bboxes. This phenomena is less prominent in fig.~\ref{fig:fig16} as it depicts the mAP for all categories combined, grouped by bounding box size.
	
	
	%To emphasize potential differences between the apertures, the results presented in Figs.~\ref{fig:fig11}-\ref{fig:fig12} are limited to tiny and small bounding boxes.
	%\clearpage
	\subsection{Hypotheses Analysis}
	By further examining Figs.~\ref{fig:fig11}-\ref{fig:fig16} a couple of observations can be made and testing their statistical significance is required:
	\begin{enumerate} 
		\item  The small variations in the mAP between the different apertures are either on the same magnitude or smaller than the standard deviation (equivalent to the ``system noise''). The null hypothesis for the aperture case states that neither of the apertures demonstrates a statistically significant advantage in average precision over the others.
		\color{black}
		In mathematical terms, the null hypothesis, $H_0$, and the alternative hypothesis, $H_a$, assume that
		\begin{eqnarray}
			H_0: \bar{\mu}_{1,i}=\bar{\mu}_{2,i}=\bar{\mu}_{3,i}=\bar{\mu}_{4,i}\\
			H_a: \bar{\mu}_{1,i}\neq\bar{\mu}_{2,i}\neq\bar{\mu}_{3,i}\neq\bar{\mu}_{4,i}
		\end{eqnarray}
		where $\bar{\mu}_{\ell,i}$ denotes the mAP measured for the $\ell$th aperture under the $i$th gain level.
		\color{black}
		\item All the apertures are fairly indifferent to additive noise incurred by camera gain of up to 40dB. A degradation was only observed in the highest gain level (of 48dB). For this case, the null hypothesis to be tested states that an increase in camera gain \color{black}that needs to be applied to images with low intensity, which amplifies the additive noise (and thus decreases the SNR)\color{black}, \emph{does not} degrade the detection precision.
		\color{black}
		In mathematical terms,
		\begin{eqnarray}
			H_0: \bar{\mu}_{\ell,1}=\bar{\mu}_{\ell,2}=\bar{\mu}_{\ell,3}=\bar{\mu}_{\ell,4}\\
			H_a: \bar{\mu}_{\ell,1}\neq\bar{\mu}_{\ell,2}\neq\bar{\mu}_{\ell,3}\neq\bar{\mu}_{\ell,4}
		\end{eqnarray}
		\color{black}
	\end{enumerate}
	
	%
	%
	%To test the null hypotheses stated above, a two-sample t-test can be applied to evaluate the t-statistic  \cite{}. 
	
	To test the null hypotheses stated above, Welch's two-sample t-test\cite{MathStats, Welch} for unequal variances may be used. In our case, as there are 4 different apertures and 4 different gain levels, the test should be conducted on all 6 pairwise combinations available with $N=4$ apertures or gain levels, calculated as $\frac{N(N-1)}{2}=6$.   
	
	For this test, the t-statistic and the degrees of freedom, $\nu$, are given by
	\begin{eqnarray}%\begin{eqnarray}
		%	se = \sqrt{((s_1^2 / K) + (s_2^2 / K)}
		%\end{eqnarray}
		t  &=& \frac{\bar{x}_1 - \bar{x}_2}{\sqrt{\frac{s_1^2}{n_1} + \frac{s_2^2}{n_2}}}\\
		\nu &=& \frac{\left(\frac{s_1^2}{n_1} + \frac{s_2^2}{n_2}\right)^2}{\frac{\left(\frac{s_1^2}{n_1}\right)^2}{n_1 - 1} + \frac{\left(\frac{s_2^2}{n_2}\right)^2}{n_2 - 1}}
	\end{eqnarray}
	where $\bar{x}_1, \bar{x}_2$ and ${s}_1, {s}_2$ correspond to the mean values and the standard deviations of each of the groups under test. In our case, the number of samples in each group equals $n_1=n_2=K (=5)$. 
	
	The two-tailed p-value is used to determine the significance of a test statistic in a hypothesis test.
	\begin{eqnarray}
		\notag
		\textrm{Two-tailed p-value}&=&2\cdot P(Z\geq|t|)\\
		&=&2\cdot(1-F_t(|t|,\nu))
	\end{eqnarray}
	where $F_t(|t|,\nu)$ denotes the cumulative distribution function of the t-distribution. 
	
	To reject the null hypothesis, $H_0$, it is required that
	\begin{eqnarray}
		2\cdot(1-F_t(|t|,\nu)) < \alpha
	\end{eqnarray}
	where commonly, $\alpha=0.05$.
	
	Notice that the null hypotheses concerning the apertures should be tested at each gain level separately. Hence, for each subplot in Figs.\ref{fig:fig11}-\ref{fig:fig16}, 6$\times$4=24 two-sample tests are required. That leads to a total of 4 gains $\times$ 6 aperture-pairs$\times$(4 figures$\times$3 subplots + 4 subplots) = 384 two-sample tests, with the gain-related null hypotheses required a similar amount of tests. For brevity, we will only present selected test examples. 
	Tables~\ref{tab:AllClassmAP_tiny_aperture}-\ref{tab:AllClassmAP_medium_gain} examine the hypotheses described above for tiny and small bbox sizes. These comparisons are based on the results presented in Fig.~\ref{fig:fig16}, in which each subplot describes the mAP for a specific bbox size target for all classes combined. As observed, tiny and small bbox sizes challenge the object detection system. From tables~\ref{tab:AllClassmAP_tiny_aperture} and\ref{tab:AllClassmAP_small_aperture} it is evident that the null hypothesis cannot be rejected for any of the cases, implying that there is no single aperture that improves the detection precision for the automotive classes considered.  
	
	With regards to the precision reduction with the increase of the gain (thereby, the decrease in SNR), it is clear from tables~\ref{tab:AllClassmAP_tiny_gain}, \ref{tab:AllClassmAP_small_gain}, and \ref{tab:AllClassmAP_medium_gain} that a gain of 48dB imposes a statistically significant mAP reduction. However, the dependence on the gain is not uniform for all apertures; some exhibit independence between the average precision and gains lower than 48 dB.         
	\begin{table}[!ht]
		\tiny
		\centering
		\begin{tabular}{ccccccc}
			\hline
			\textbf{Gain [dB]} & \textbf{Aperture} \#1 & \textbf{Aperture} \#2 &  \textbf{t-statistic} & $\boldsymbol{\nu}$ & \textbf{p-value} & \makecell{\textbf{Reject} $H_0$\\ \textbf{(p < 0.05)?}} \\
			\hline
			0dB & Plus F/2.4 & Horizontal F/3.4 & 0.2848 & 7.6672 & 0.7834 & no \\
			0dB & Plus F/2.4 & Vertical F/3.4 & 0.3919 & 7.5521 & 0.7060 & no \\
			0dB & Plus F/2.4 & Circular F/1.8 & 0.4274 & 7.7926 & 0.6806 & no \\
			0dB & Horizontal F/3.4 & Vertical F/3.4 & 0.0304 & 6.7513 & 0.9766 & no \\
			0dB & Horizontal F/3.4 & Circular F/1.8 & 0.1219 & 7.9825 & 0.9060 & no \\
			0dB & Vertical F/3.4 & Circular F/1.8 & 0.1159 & 6.9385 & 0.9110 & no \\
			\hline
			30dB & Plus F/2.4 & Horizontal F/3.4 & -1.6044 & 7.1096 & 0.1520 & no \\
			30dB & Plus F/2.4 & Vertical F/3.4 & -0.5266 & 7.9963 & 0.6127 & no \\
			30dB & Plus F/2.4 & Circular F/1.8 & -0.4961 & 7.9963 & 0.6332 & no \\
			30dB & Horizontal F/3.4 & Vertical F/3.4 & 1.1936 & 7.0254 & 0.2714 & no \\
			30dB & Horizontal F/3.4 & Circular F/1.8 & 1.2181 & 7.0254 & 0.2625 & no \\
			30dB & Vertical F/3.4 & Circular F/1.8 & 0.0309 & 8.0000 & 0.9761 & no \\
			\hline
			40dB & Plus F/2.4 & Horizontal F/3.4 & -0.1969 & 6.7269 & 0.8497 & no \\
			40dB & Plus F/2.4 & Vertical F/3.4 & 1.1151 & 7.8663 & 0.2977 & no \\
			40dB & Plus F/2.4 & Circular F/1.8 & -0.3066 & 7.9295 & 0.7671 & no \\
			40dB & Horizontal F/3.4 & Vertical F/3.4 & 1.0190 & 6.2195 & 0.3462 & no \\
			40dB & Horizontal F/3.4 & Circular F/1.8 & -0.0439 & 7.1034 & 0.9662 & no \\
			40dB & Vertical F/3.4 & Circular F/1.8 & -1.3787 & 7.6244 & 0.2071 & no \\
			\hline
			48dB & Plus F/2.4 & Horizontal F/3.4 & -0.9735 & 7.5163 & 0.3606 & no \\
			48dB & Plus F/2.4 & Vertical F/3.4 & -1.0168 & 7.9894 & 0.3390 & no \\
			48dB & Plus F/2.4 & Circular F/1.8 & -0.8566 & 6.3894 & 0.4226 & no \\
			48dB & Horizontal F/3.4 & Vertical F/3.4 & 0.0775 & 7.6331 & 0.9402 & no \\
			48dB & Horizontal F/3.4 & Circular F/1.8 & 0.4268 & 5.5196 & 0.6857 & no \\
			48dB & Vertical F/3.4 & Circular F/1.8 & 0.4017 & 6.2515 & 0.7013 & no \\
			\hline
		\end{tabular}
		\caption{All Class mAP, Aperture Significance: Tiny-sized bbox < 23x23 pixels}
		\label{tab:AllClassmAP_tiny_aperture}
	\end{table}
	\begin{table}[!ht]
		\tiny
		\centering
		\begin{tabular}{ccccccc}
			\hline
			\textbf{Gain [dB]} & \textbf{Aperture} \#1 & \textbf{Aperture} \#2 &  \textbf{t-statistic} & $\boldsymbol{\nu}$ & \textbf{p-value} & \makecell{\textbf{Reject} $H_0$\\ \textbf{(p < 0.05)?}} \\
			\hline
			0dB & Plus F/2.4 & Horizontal F/3.4 & -1.5737 & 6.5452 & 0.1625 & no \\
			0dB & Plus F/2.4 & Vertical F/3.4 & -0.8101 & 6.5452 & 0.4463 & no \\
			0dB & Plus F/2.4 & Circular F/1.8 & -1.1545 & 7.9928 & 0.2817 & no \\
			0dB & Horizontal F/3.4 & Vertical F/3.4 & 1.0503 & 8.0000 & 0.3243 & no \\
			0dB & Horizontal F/3.4 & Circular F/1.8 & 0.1981 & 6.6639 & 0.8488 & no \\
			0dB & Vertical F/3.4 & Circular F/1.8 & -0.5824 & 6.6639 & 0.5795 & no \\
			\hline
			30dB & Plus F/2.4 & Horizontal F/3.4 & -1.6997 & 7.6803 & 0.1292 & no \\
			30dB & Plus F/2.4 & Vertical F/3.4 & -1.1901 & 7.3409 & 0.2711 & no \\
			30dB & Plus F/2.4 & Circular F/1.8 & -1.0324 & 7.9284 & 0.3323 & no \\
			30dB & Horizontal F/3.4 & Vertical F/3.4 & 0.6521 & 7.9178 & 0.5328 & no \\
			30dB & Horizontal F/3.4 & Circular F/1.8 & 0.6521 & 7.9024 & 0.5329 & no \\
			30dB & Vertical F/3.4 & Circular F/1.8 & 0.0693 & 7.6602 & 0.9465 & no \\
			\hline
			40dB & Plus F/2.4 & Horizontal F/3.4 & -1.2575 & 7.8088 & 0.2449 & no \\
			40dB & Plus F/2.4 & Vertical F/3.4 & -0.3967 & 4.2166 & 0.7109 & no \\
			40dB & Plus F/2.4 & Circular F/1.8 & -1.0589 & 7.0800 & 0.3244 & no \\
			40dB & Horizontal F/3.4 & Vertical F/3.4 & 1.4391 & 4.2967 & 0.2188 & no \\
			40dB & Horizontal F/3.4 & Circular F/1.8 & 0.3376 & 7.6428 & 0.7447 & no \\
			40dB & Vertical F/3.4 & Circular F/1.8 & -1.2507 & 4.4595 & 0.2726 & no \\
			\hline
			48dB & Plus F/2.4 & Horizontal F/3.4 & -1.4329 & 7.0540 & 0.1947 & no \\
			48dB & Plus F/2.4 & Vertical F/3.4 & -1.6996 & 7.8977 & 0.1281 & no \\
			48dB & Plus F/2.4 & Circular F/1.8 & -0.5710 & 7.1542 & 0.5855 & no \\
			48dB & Horizontal F/3.4 & Vertical F/3.4 & -0.0042 & 7.4810 & 0.9967 & no \\
			48dB & Horizontal F/3.4 & Circular F/1.8 & 1.1370 & 5.7236 & 0.3009 & no \\
			48dB & Vertical F/3.4 & Circular F/1.8 & 1.4055 & 6.7004 & 0.2045 & no \\
			\hline
		\end{tabular}
		\caption{All Class mAP, Aperture Significance:  Small-sized bbox, 23$\times$23 < bbox < 32$\times$32 pixels}
		\label{tab:AllClassmAP_small_aperture}
	\end{table}

	\begin{table}[ht]
		\centering
		\tiny
		\begin{tabular}{ccccccc}
			\hline
			\textbf{Gain [dB]} & \textbf{Aperture} \#1 & \textbf{Aperture} \#2 &  \textbf{t-statistic} & $\boldsymbol{\nu}$ & \textbf{p-value} & \makecell{\textbf{Reject} $H_0$\\ \textbf{(p < 0.05)?}} \\
			\hline
			Plus F/2.4 & 0dB & 30dB & 0.9938 & 7.0268 & 0.3533 & no \\
			Plus F/2.4 & 0dB & 40dB & 0.9852 & 7.2231 & 0.3564 & no \\
			Plus F/2.4 & 0dB & 48dB & 7.2723 & 7.8474 & 0.0001 & \textbf{yes} \\
			Plus F/2.4 & 30dB & 40dB & 0.0095 & 7.9798 & 0.9927 & no \\
			Plus F/2.4 & 30dB & 48dB & 7.4070 & 6.4695 & 0.0002 & \textbf{yes} \\
			Plus F/2.4 & 40dB & 48dB & 7.3014 & 6.6678 & 0.0002 & \textbf{yes} \\
			\hline
			Horizontal F/3.4 & 0dB & 30dB & -0.7372 & 7.6034 & 0.4831 & no \\
			Horizontal F/3.4 & 0dB & 40dB & 0.2940 & 7.9417 & 0.7763 & no \\
			Horizontal F/3.4 & 0dB & 48dB & 4.5441 & 7.7323 & 0.0021 & \textbf{yes} \\
			Horizontal F/3.4 & 30dB & 40dB & 1.1048 & 7.8341 & 0.3020 & no \\
			Horizontal F/3.4 & 30dB & 48dB & 5.5840 & 6.9082 & 0.0009 & \textbf{yes} \\
			Horizontal F/3.4 & 40dB & 48dB & 4.4344 & 7.4658 & 0.0026 & \textbf{yes} \\
			\hline
			Vertical F/3.4 & 0dB & 30dB & 0.1997 & 7.7946 & 0.8468 & no \\
			Vertical F/3.4 & 0dB & 40dB & 1.7694 & 7.6323 & 0.1166 & no \\
			Vertical F/3.4 & 0dB & 48dB & 6.2453 & 6.8896 & 0.0005 & \textbf{yes} \\
			Vertical F/3.4 & 30dB & 40dB & 1.7179 & 7.9720 & 0.1243 & no \\
			Vertical F/3.4 & 30dB & 48dB & 6.3742 & 6.2492 & 0.0006 & \textbf{yes} \\
			Vertical F/3.4 & 40dB & 48dB & 5.2993 & 6.0347 & 0.0018 & \textbf{yes} \\
			\hline
			Circular F/1.8 & 0dB & 30dB & 0.0514 & 6.2951 & 0.9606 & no \\
			Circular F/1.8 & 0dB & 40dB & 0.1585 & 6.9489 & 0.8786 & no \\
			Circular F/1.8 & 0dB & 48dB & 6.8686 & 6.2974 & 0.0004 & \textbf{yes} \\
			Circular F/1.8 & 30dB & 40dB & 0.1511 & 7.7883 & 0.8838 & no \\
			Circular F/1.8 & 30dB & 48dB & 9.8426 & 8.0000 & 0.0000 & \textbf{yes} \\
			Circular F/1.8 & 40dB & 48dB & 8.8442 & 7.7898 & 0.0000 & \textbf{yes} \\
			\hline
		\end{tabular}
		\caption{All Class mAP, Gain Significance: Tiny-sized bbox < 23$x$23 pixels}
		\label{tab:AllClassmAP_tiny_gain}
	\end{table}

	\begin{table}[!ht]
		\centering
		\tiny
		\begin{tabular}{ccccccc}
			\hline
			\textbf{Gain [dB]} & \textbf{Aperture} \#1 & \textbf{Aperture} \#2 &  \textbf{t-statistic} & $\boldsymbol{\nu}$ & \textbf{p-value} & \makecell{\textbf{Reject} $H_0$\\ \textbf{(p < 0.05)?}} \\
			\hline
			Plus F/2.4 & 0dB & 30dB & 0.4424 & 7.9140 & 0.6700 & no \\
			Plus F/2.4 & 0dB & 40dB & 1.4219 & 7.9949 & 0.1929 & no \\
			Plus F/2.4 & 0dB & 48dB & 9.2920 & 7.6045 & 0.0000 & \textbf{yes} \\
			Plus F/2.4 & 30dB & 40dB & 1.0560 & 7.8687 & 0.3223 & no \\
			Plus F/2.4 & 30dB & 48dB & 9.3865 & 7.8734 & 0.0000 & \textbf{yes} \\
			Plus F/2.4 & 40dB & 48dB & 7.5764 & 7.5228 & 0.0001 & \textbf{yes} \\
			\hline
			Horizontal F/3.4 & 0dB & 30dB & 0.4831 & 7.6990 & 0.6425 & no \\
			Horizontal F/3.4 & 0dB & 40dB & 2.0498 & 7.0721 & 0.0791 & no \\
			Horizontal F/3.4 & 0dB & 48dB & 8.9169 & 5.9814 & 0.0001 & \textbf{yes} \\
			Horizontal F/3.4 & 30dB & 40dB & 1.5051 & 7.7565 & 0.1719 & no \\
			Horizontal F/3.4 & 30dB & 48dB & 8.1568 & 6.7373 & 0.0001 & \textbf{yes} \\
			Horizontal F/3.4 & 40dB & 48dB & 6.5207 & 7.4299 & 0.0003 & \textbf{yes} \\
			\hline
			Vertical F/3.4 & 0dB & 30dB & 0.2357 & 7.9241 & 0.8197 & no \\
			Vertical F/3.4 & 0dB & 40dB & 4.1259 & 4.6307 & 0.0107 & \textbf{yes} \\
			Vertical F/3.4 & 0dB & 48dB & 10.0543 & 7.0142 & 0.0000 & \textbf{yes} \\
			Vertical F/3.4 & 30dB & 40dB & 3.4567 & 4.5193 & 0.0213 & \textbf{yes} \\
			Vertical F/3.4 & 30dB & 48dB & 9.5400 & 7.3889 & 0.0000 & \textbf{yes} \\
			Vertical F/3.4 & 40dB & 48dB & 9.0739 & 4.2882 & 0.0006 & \textbf{yes} \\
			\hline
			Circular F/1.8 & 0dB & 30dB & 0.7462 & 7.7797 & 0.4775 & no \\
			Circular F/1.8 & 0dB & 40dB & 1.9433 & 7.2927 & 0.0914 & no \\
			Circular F/1.8 & 0dB & 48dB & 11.5577 & 6.3571 & 0.0000 & \textbf{yes} \\
			Circular F/1.8 & 30dB & 40dB & 1.2799 & 7.8215 & 0.2372 & no \\
			Circular F/1.8 & 30dB & 48dB & 12.1053 & 7.0280 & 0.0000 & \textbf{yes} \\
			Circular F/1.8 & 40dB & 48dB & 11.8253 & 7.5848 & 0.0000 & \textbf{yes} \\
			\hline
		\end{tabular}
		\caption{All Class mAP, Aperture Significance: Small-sized bbox, 23$x$23 < bbox < 32$x$32 pixels}
		\label{tab:AllClassmAP_small_gain}
	\end{table}
	
	\begin{table}[!ht]
		\centering
		\tiny
		\begin{tabular}{ccccccc}
			\hline
			\textbf{Gain [dB]} & \textbf{Aperture} \#1 & \textbf{Aperture} \#2 &  \textbf{t-statistic} & $\boldsymbol{\nu}$ & \textbf{p-value} & \makecell{\textbf{Reject} $H_0$\\ \textbf{(p < 0.05)?}} \\
			\hline
			Plus F/2.4 & 0dB & 30dB & 1.6421 & 7.6536 & 0.1409 & no \\
			Plus F/2.4 & 0dB & 40dB & 5.1541 & 6.1527 & 0.0020 & \textbf{yes} \\
			Plus F/2.4 & 0dB & 48dB & 19.5159 & 7.9837 & 0.0000 & \textbf{yes} \\
			Plus F/2.4 & 30dB & 40dB & 2.3944 & 5.4640 & 0.0578 & no \\
			Plus F/2.4 & 30dB & 48dB & 15.5623 & 7.5096 & 0.0000 & \textbf{yes} \\
			Plus F/2.4 & 40dB & 48dB & 19.2516 & 6.3206 & 0.0000 & \textbf{yes} \\
			\hline
			Horizontal F/3.4 & 0dB & 30dB & 0.2226 & 7.4079 & 0.8299 & no \\
			Horizontal F/3.4 & 0dB & 40dB & 1.3075 & 7.4340 & 0.2300 & no \\
			Horizontal F/3.4 & 0dB & 48dB & 15.3165 & 6.9699 & 0.0000 & \textbf{yes} \\
			Horizontal F/3.4 & 30dB & 40dB & 0.9561 & 7.9996 & 0.3670 & no \\
			Horizontal F/3.4 & 30dB & 48dB & 12.0711 & 5.8733 & 0.0000 & \textbf{yes} \\
			Horizontal F/3.4 & 40dB & 48dB & 10.9295 & 5.8977 & 0.0000 & \textbf{yes} \\
			\hline
			Vertical F/3.4 & 0dB & 30dB & 0.6666 & 7.9555 & 0.5239 & no \\
			Vertical F/3.4 & 0dB & 40dB & 1.2189 & 7.2687 & 0.2609 & no \\
			Vertical F/3.4 & 0dB & 48dB & 14.7776 & 7.1456 & 0.0000 & \textbf{yes} \\
			Vertical F/3.4 & 30dB & 40dB & 0.7045 & 6.9771 & 0.5039 & no \\
			Vertical F/3.4 & 30dB & 48dB & 14.5934 & 6.8489 & 0.0000 & \textbf{yes} \\
			Vertical F/3.4 & 40dB & 48dB & 11.8964 & 7.9918 & 0.0000 & \textbf{yes} \\
			\hline
			Circular F/1.8 & 0dB & 30dB & 0.4581 & 7.9267 & 0.6591 & no \\
			Circular F/1.8 & 0dB & 40dB & 2.8853 & 7.6639 & 0.0213 & \textbf{yes} \\
			Circular F/1.8 & 0dB & 48dB & 16.7842 & 7.5053 & 0.0000 & \textbf{yes} \\
			Circular F/1.8 & 30dB & 40dB & 2.5427 & 7.8945 & 0.0349 & \textbf{yes} \\
			Circular F/1.8 & 30dB & 48dB & 17.2735 & 7.7888 & 0.0000 & \textbf{yes} \\
			Circular F/1.8 & 40dB & 48dB & 15.6500 & 7.9800 & 0.0000 & \textbf{yes} \\
			\hline
		\end{tabular}
		\caption{All Class mAP, Aperture Significance: Medium-sized bbox, 32x32 < bbox < 96x96 pixels}
		\label{tab:AllClassmAP_medium_gain}
	\end{table}
	
	%\clearpage	
	\section{Summary \& Conclusions}
	In this paper, we presented a deep learning-based comparative numerical analysis of automotive object detection tasks, examining four different aperture types with varying shapes and cross-sections at different camera gain levels. Our gain-related analysis showed a weak dependence of detection precision on additive noise levels across all aperture shapes, up to a critical noise threshold.
	
	Regarding the relationship between detection precision and aperture characteristics, the findings are twofold: first, no statistically significant difference in detection precision was observed between different aperture shapes. Second, the results indicate no significant difference in detection performance across f-numbers ranging from $f/1.8$ to $f/3.4$.
	The observation that higher f-number values do not degrade object detection performance suggests a potential cost-saving opportunity in lens specifications, as optics with smaller diameters tend to be more economical.
	\color{black}Furthermore, recall that the minimal spot size of an optical system is defined as the resolution limit for a diffraction-limited system. Such a spot should typically encompass four adjacent pixels to effectively extract a feature. Assuming a wavelength of $\lambda$ (specified in $[\mu m]$) the minimal spot size is defined as \cite{AiryDisk}
	\begin{eqnarray}\label{eq:spotsize}
		\textrm{Spot Size}\approx4~\textrm{pixels}=2\cdot1.22\cdot\frac{\lambda}{NA}=4\cdot1.22\cdot\lambda\cdot f/\#
	\end{eqnarray}  
	where the factor 1.22 originates from the calculation of the position of the first dark circular ring surrounding the central Airy disk in the diffraction pattern, which corresponds to the first zero of the order-one Bessel function of the 
	first kind $J_1(x)$ \cite[Table 4.1, p.78]{AiryDisk}, \color{black} and $NA$ denotes the ``numerical aperture'' of the lens 
	\begin{eqnarray}
		NA=\frac{D}{2f}=\frac{1}{2\cdot f/\#}
	\end{eqnarray}
	\noindent 
	\color{black}
	Equation \eqref{eq:spotsize} suggests that pixel size should scale proportionally with the f-number. At higher f-numbers, smaller pixels offer limited resolution benefits as the system becomes diffraction-limited. Additionally, smaller pixels feature reduced well-capacity, limited dynamic range, and increased susceptibility to noise. Maintaining the same field of view with smaller pixels requires more pixels, leading to higher memory usage, increased data rates, and greater readout demands. These factors raise system costs, offsetting any savings achieved through simpler optics.
	\color{black}

	%Therefore, a larger f-number indicates that smaller pixels do not significantly enhance camera performance as the imaging sensor may become diffraction-limited. Since high-resolution imaging sensors often balance increased pixel count with reduced pixel size, this suggests that higher resolution sensors may not always provide better performance. Additionally, larger pixels offer a higher dynamic range and are less susceptible to additive noise, which collectively contributes to improved image quality.
%	\section*{Code and Data Availability}
%	The data that support the findings of this article is owned by General Motors and cannot be made publicly available.
%	\section*{Disclosures}
%	The authors declare no conflicts of interest. 
%	\section*{Acknowledgments}
%	This research was carried out under General Motors R\&D Division. 
%	\\
%	AI tools such as Grammarly for cleaning up language and grammar have been used during the writing of this manuscript.  
%	%	\section*{Acknowledgments}
%	%	This was was supported in part by......
%	\newpage
	\appendix
	\section{Determining Object Distance}
	\label{sec:depth_calc}
	The object distances listed in Table~\ref{tab:bbox_range} were determined as follows.	
		\begin{table}[!ht]
		\centering
		\begin{tabular}{lc}
			\midrule
			\textbf{Object Type}    & \textbf{Size [cm]} \\
			\midrule
			Traffic Light & 30.5	  \\
			Traffic Sign  & 62      \\
			Speed Sign    & 25​     \\
			\bottomrule
		\end{tabular}
		\caption{Physical Object Sizes}
		\label{tab:Object_size}
	\end{table}
	If we denote the horizontal image resolution as $R$ (specified in pixels), the object width a $W$ (specified in Table\ref{tab:Object_size}), located at a distance $D$ from the focal point (where both $W$ and $D$ have the same units - commonly meters), then the width of the object's bounding box projected on the imaging plane as $w$ (specified in pixels)  is given by
	\begin{eqnarray}\label{eq:bboxwidth}
		w=\Bigg\lceil
		\frac{2\cdot\tan^{-1}\left(\frac{W}{2\cdot D}\right)}{\textrm{FOV}_{\textrm{H}}}\cdot R
		\Bigg\rceil
	\end{eqnarray}
	where $\textrm{FOV}_{\textrm{H}}$ is given by \eqref{eq:FOV_H}.
	
\begin{figure}[!ht]
		\centering
		%		\fbox{\rule[-.5cm]{4cm}{4cm} \rule[-.5cm]{4cm}{0cm}}
		\resizebox{3.5in}{!}{\includegraphics{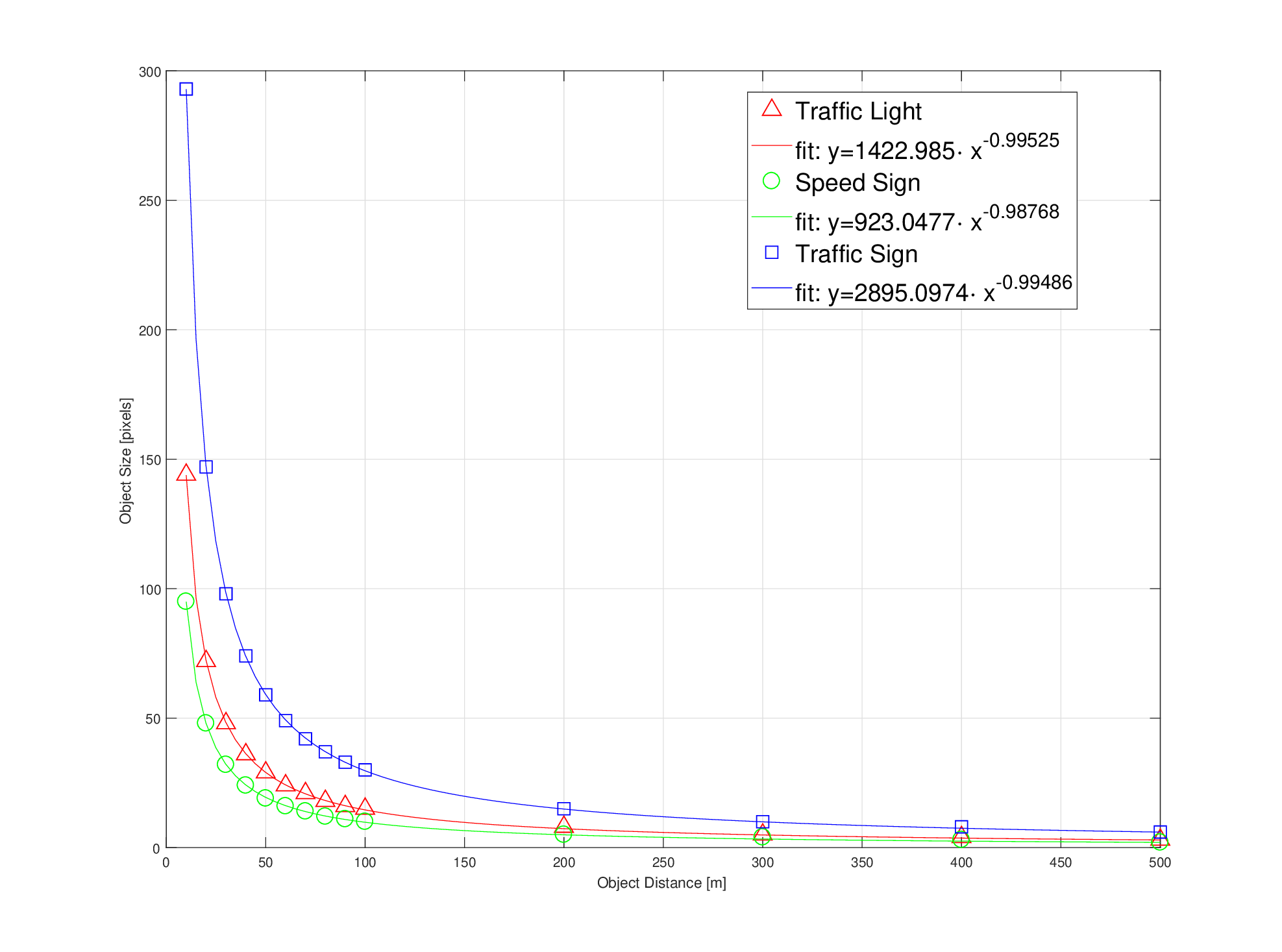}}
		\caption{Object Bounding Box Width vs. Object Distance}
		\label{fig:fig26}
	\end{figure}

	By applying \eqref{eq:bboxwidth}, Fig.\ref{fig:fig26} illustrates the relationship between bounding box width and object type as a function of distance. Through power-law curve fitting, specific bounding box widths can be mapped to their corresponding distances, as detailed in Table~\ref{tab:bbox_range}.

	\begin{IEEEbiography}[{\includegraphics[width=1in,height=1.25in,clip,keepaspectratio]{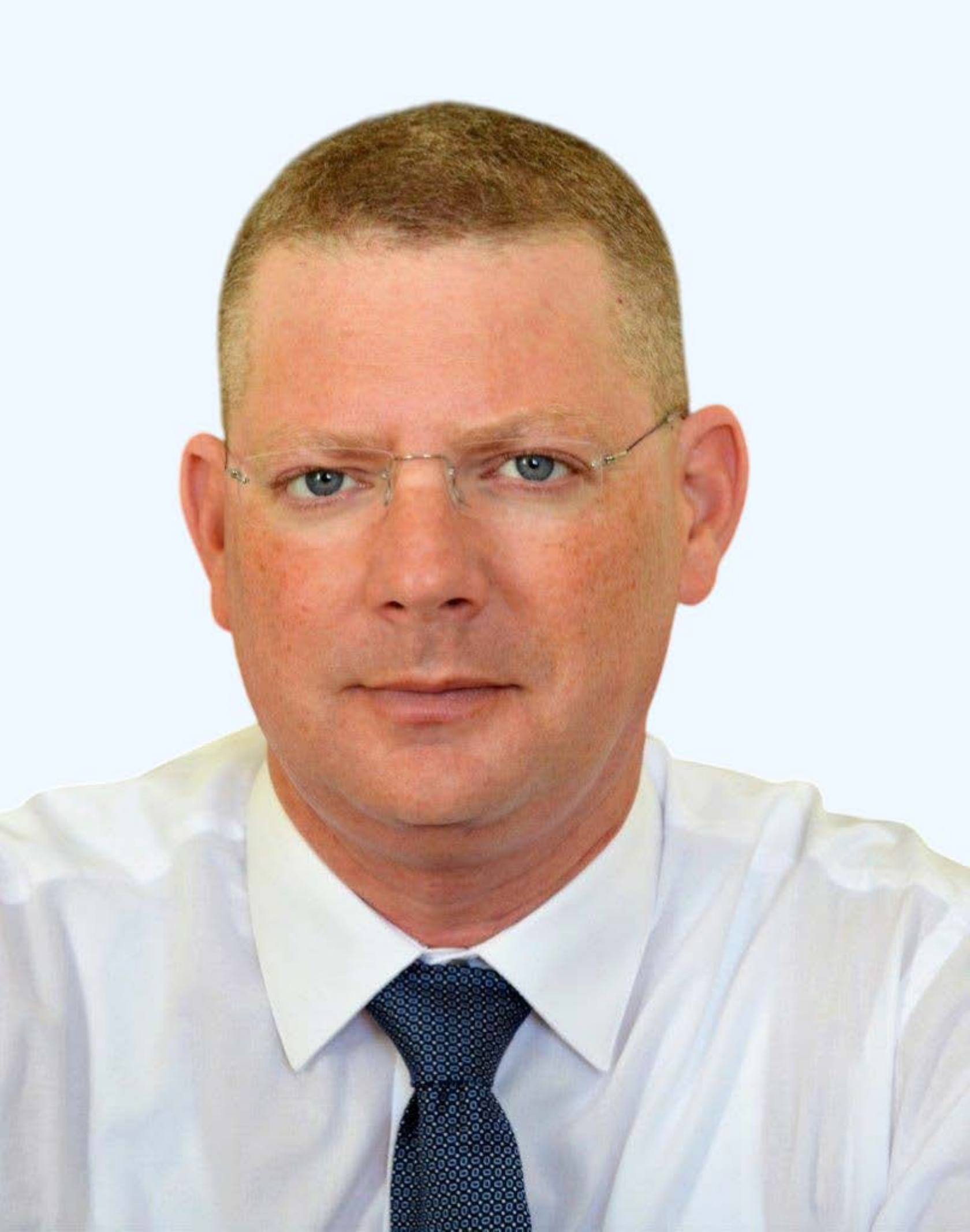}}]{Ofer Bar-Shalom}
received the B.Sc. degree in mechanical engineering and the M.Sc. and Ph.D. 
degrees in electrical engineering from Tel-Aviv University, Tel Aviv, Israel, in 1997, 2001, and 2015,  respectively. He has been involved in the development of cellular and wireless connectivity systems for over 20 years. He is currently a Senior Researcher with General Motors R\&D Division, Herzliya, Israel. He has authored multiple journal and conference papers and holds over 20 patents in wireless communications, real-time systems, multimedia, and geolocation applications. His research interests include signal processing, computer vision, geolocation, navigation and radar systems, and deep learning applications.
	\end{IEEEbiography}
	\vfill
	\begin{IEEEbiography}[{\includegraphics[width=1in,height=1.25in,clip,keepaspectratio]{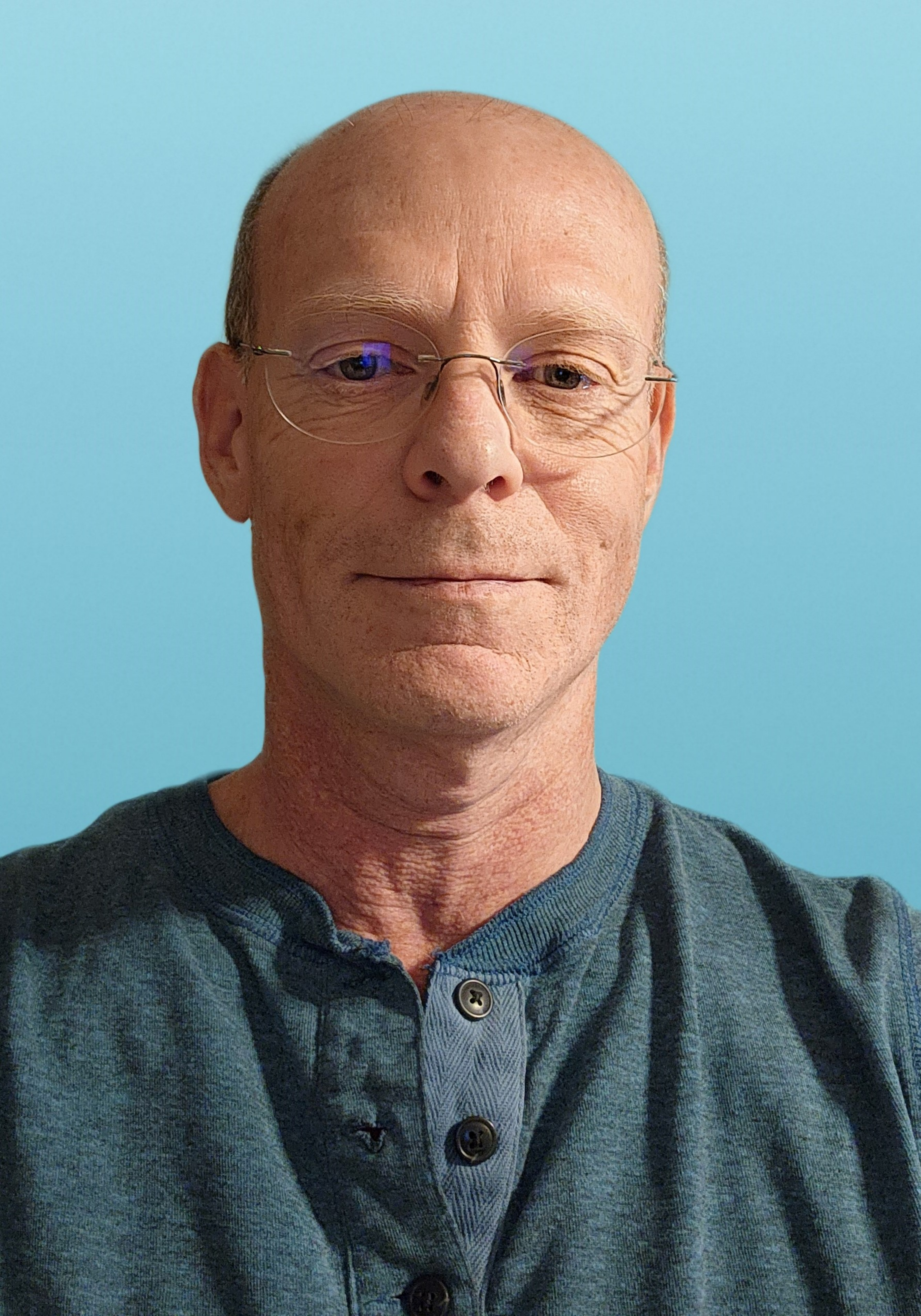}}]{Tzvi Philipp}
	received his BA in physics from Yeshiva University in 1988, his B.Sc. in applied physics from Columbia University in 1990, and his MS in physics from The City University of New York in 1992. He has led optical design projects across the medical, semiconductors and automotive industries for over 30 years. He is currently a Staff Researcher with General Motors R\&D division focused on developing smart sensing and vision systems. His research interests include efficient optical designs, image and video processing, and computer vision.  
	\end{IEEEbiography}
	\begin{IEEEbiography}[{\includegraphics[width=1in,height=1.25in,clip,keepaspectratio]{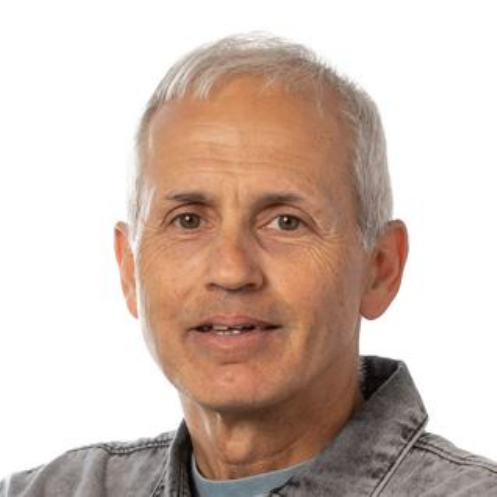}}]{Eran Kishon}
		received his B.Sc. and M.Sc. degrees in Electrical Engineering from Tel-Aviv University, Tel Aviv, Israel, in 1987 and 2003, respectively. Lead architecture and algorithm development in image and video-related consumer and industrial applications.
		Currently a Staff Researcher with General Motors R\&D Division, Herzliya, Israel.
		His research interests include image and video processing, computer vision, machine and deep learning, and generative models.
	\end{IEEEbiography}
	\vfill
	%\newpage

	
	%\vfill
	


\begin{thebibliography}{100}
		\bibitem{CMOSsensors}
		A.~El Gamal and H.~Eltoukhy, ``CMOS Image Sensors,'' \textit{IEEE Circuits \& Devices}, Magazine May/June 2005, pp.~6--20.
		\bibitem{Yadid}
		O. Yadid-Pecht and R. Etienne-Cummings (Eds.), \textit{CMOS Imagers: From Phototransduction to Image Processing},  Kluwer Academic Publishers, 2004.
		
		\bibitem{Banks}
		M. S. Banks, W. W. Sprague, J. Schmoll, J. A. Q. Parnell and G. D. Love, ``Why do animal eyes have pupils of different shapes?'' \textit{Science Advance}, Aug 7, 2015;1(7):e1500391. doi: [online:]~\url{https://www.science.org/doi/10.1126/sciadv.1500391}. PMID: 26601232; PMCID: PMC4643806.
		
		
		\bibitem{YOLOorig}
		J. Redmon, S. Divvala, R. Girshick, A. Farhadi, ``You Only Look Once: Unified, Real-Time Object Detection''
		[online:] \url{https://arxiv.org/abs/1506.02640}
		\bibitem{YOLOv8}
		[online:]~\url{https://docs.ultralytics.com/models/yolov8/}
		\bibitem{bdd}
		[online:]~\url{http://bdd-data.berkeley.edu/}
		\bibitem{kitti}
		[online:]~\url{https://www.cvlibs.net/datasets/kitti/}
		\bibitem{CognataTS}
		[online:]~\url{https://www.cognata.com/traffic-sign-datasets/}
		\bibitem{CognataTL}
		[online:]~\url{https://www.cognata.com/traffic-lights/}
		\bibitem{coco}
		T.-Y. Lin, M. Maire, S. Belongie, L. Bourdev, R. Girshic, J. Hays, P. Perona, D. Ramanan, C. L. Zitnick and P. Dollár, ``Microsoft COCO: Common Objects in Context,''
		[online]~\url{https://arxiv.org/pdf/1405.0312}
		\bibitem{Camera}
		\textit{DFK 37BUX252 Technical Reference Manual}, [online:]~\url{https://www.theimagingsource.com/en-us/product/industrial/37u/dfk37bux252/}
		
		\bibitem{Tseng2021DeepCompoundOptics}
		E. Tseng, A. Mosleh, F. Mannan, K. St-Arnaud, A. Sharma, And Y. Peng, A. Braun, D. Nowrouzezahrai, J.-F. Lalonde And F. Heide, ``Supplementary Information Differentiable Compound Optics and Processing Pipeline Optimization for End-to-end Camera Design,'' \textit{ACM Transactions on Graphics (TOG)}, Vol. 40, No. 2, Article 18, August 2021.
		
		\bibitem{HWinLoop}
		A. Mosleh, A. Sharma, E. Onzon, F. Mannan, N. Robidoux and F. Heide, ``Hardware-in-the-Loop End-to-End Optimization of Camera Image Processing Pipelines,'' \textit{2020 IEEE/CVF Conference on Computer Vision and Pattern Recognition (CVPR)}, Seattle, WA, USA, 2020, pp. 7526-7535, ~[online:]~\url{https://ieeexplore.ieee.org/document/9156332}.
		
		\bibitem{DiffLens}
		G. C{\^o}t{\'e}, F. Mannan, S. Thibault, J.-F. Lalonde and F. Heide, ``The Differentiable Lens: Compound Lens Search over Glass Surfaces and Materials for Object Detection,'' \textit{Proc. of the IEEE/CVF Conference on Computer Vision and Pattern Recognition (CVPR)}, June $18^{\textrm{th}}$-$22^{\textrm{nd}}$, 2023.
		\bibitem{SPIE_PSF}
		J. Carney, H. Corbett, W. Marshall, N. Law, S. Fitton, R. Gonzalez, L. Machia, T. Proctor, and A. Vasquez, ``PSF modeling with deep learning for the Argus Optical Array,'' \textit{Proc. SPIE 13101, Software and Cyberinfrastructure for Astronomy VIII}, 131010O (26 July 2024); [online:]~\url{https://doi.org/10.1117/12.3020465}
		
		\bibitem{MonoDepth}
		J. Chang and G. Wetzstein, ``Deep Optics for Monocular Depth Estimation and 3D Object Detection,''~[online:]~\url{https://arxiv.org/abs/1904.08601}
		
		\bibitem{FastPSF}
		J. Herbel, T. Kacprzak,a A. Amara A. Refregiera and A. Lucchib, ``Fast point spread function modeling with deep learning,'' Journal of Cosmology and Astroparticle Physics An IOP and SISSA journal, Vol. 54, July 2018,~[online:]~\url{https://doi.org/10.1088/1475-7516/2018/07/054}. 
		
		\bibitem{WolfWindshield}
		D. W. Wolf, M. Ulrich and N. Kapoor, ``Sensitivity analysis of AI-based algorithms for autonomous driving on optical wavefront aberrations induced by the windshield,'' 2023 IEEE/CVF International Conference on Computer Vision Workshops (ICCVW), Paris, France, 2023, pp. 4102-4111, [online:~\url{https://arxiv.org/abs/2308.11711}]
		
		\bibitem{Kfold}
		P. Refaeilzadeh, L. Tang, H. Liu, (2009). Cross-Validation. In: L. LIU , M.T. Özsu, (eds) Encyclopedia of Database Systems. Springer, Boston, MA. \url{https://doi.org/10.1007/978-0-387-39940-9_565}
		
		\bibitem{AiryDisk}
		J. W. Goodman, \textit{Introduction to Fourier Optics,} 3rd ed., Roberts \& Company Publishers, 2005.
		%G. B. Airy, ``On the Diffraction of an Object-glass with Circular Aperture,'' \textit{Transactions of the Cambridge Philosophical Society}, vol. 5, pp. 283–91, 1835.  [online:]~\url{https://archive.org/details/transactionsofca05camb/page/286/mode/2up}
		
		\bibitem{Dataplot}
		N. A. Heckert and James J. Filliben ``NIST Handbook 148: DATAPLOT Reference Manual, Volume I: Commands,'' \textit{National Institute of Standards and Technology Handbook Series}, June 2003.
		[online:]~\url{https://www.itl.nist.gov/div898/software/dataplot/refman2/ch2/weightsd.pdf}
		
		\bibitem{MathStats}
		R. J. Larsen and M. L. Marx, \textit{An Introduction to Mathematical Statistics and Its Applications,} Pearson-Prentice Hall Upper Saddle River, New Jersey,  07458, 4th Ed. 2006. 
		
		\bibitem{Welch}
		B. Derrick and P. White, ``Why Welch's test is Type I error robust,'' \textit{The Quantitative Methods for Psychology}, vol. 12, no. 1, pp. 30--38, 2016 [online:]~\url{http://www.tqmp.org/RegularArticles/vol12-1/p030/p030.pdf}.
		
		\bibitem{Szeliski}
		R. Szeliski, \textit{Computer Vision Algorithms and Applications,} Springer-Verlag London, 2011.
	\end{thebibliography}
\end{document}